\documentclass[10pt,twocolumn,letterpaper]{article}

\usepackage{iccv}
\usepackage{times}
\usepackage{epsfig}
\usepackage{graphicx}
\usepackage{amsmath}
\usepackage{amssymb}
\usepackage{tabularx}
\usepackage{cite}
\usepackage{authblk}
\usepackage{multicol}
\usepackage{threeparttable}
\usepackage{subcaption}

\usepackage[breaklinks=true,bookmarks=false]{hyperref}

\iccvfinalcopy 

\def\iccvPaperID{****} 

\begin{document}

\title{Local Supports Global:\\ Deep Camera Relocalization with Sequence Enhancement}

\author[1,2]{Fei Xue} 
\author[2,3]{Xin Wang}
\author[2,3]{Zike Yan}
\author[2,3]{Qiuyuan Wang} 
\author[4]{Junqiu Wang}
\author[2,3]{Hongbin Zha}
\affil[1]{UISEE Technology Inc.}
\affil[2]{Key Laboratory of Machine Perception, Peking University}
\affil[3] {PKU-SenseTime Machine Vision Joint Lab, Peking University}
\affil[4]{Beijing Changcheng Aviation Measurement and Control Institute}
\affil[ ]{\tt\small \{feixue, xinwang\_cis, zike.yan, wangqiuyuan\}@pku.edu.cn \authorcr
	\tt\small jerywangjq@foxmail.com, zha@cis.pku.edu.cn}

\maketitle

\begin{abstract}
	We propose to leverage the local information in image sequences to support global camera relocalization. In contrast to previous methods that regress global poses from single images, we exploit the spatial-temporal consistency in sequential images to alleviate uncertainty due to visual ambiguities by incorporating a visual odometry (VO) component. Specifically, we introduce two effective steps called content-augmented pose estimation and motion-based refinement. The content-augmentation step focuses on alleviating the uncertainty of pose estimation by augmenting the observation based on the co-visibility in local maps built by the VO stream. Besides, the motion-based refinement is formulated as a pose graph, where the camera poses are further optimized by adopting relative poses provided by the VO component as additional motion constraints. Thus, the global consistency can be guaranteed. Experiments on the public indoor 7-Scenes and outdoor Oxford RobotCar benchmark datasets demonstrate that benefited from local information inherent in the sequence, our approach outperforms state-of-the-art methods, especially in some challenging cases, e.g., insufficient texture, highly repetitive textures, similar appearances, and over-exposure. 

	
\end{abstract}

\section{Introduction}
\label{introduction}

\begin{figure}[t]
	\includegraphics[width=1.\linewidth]{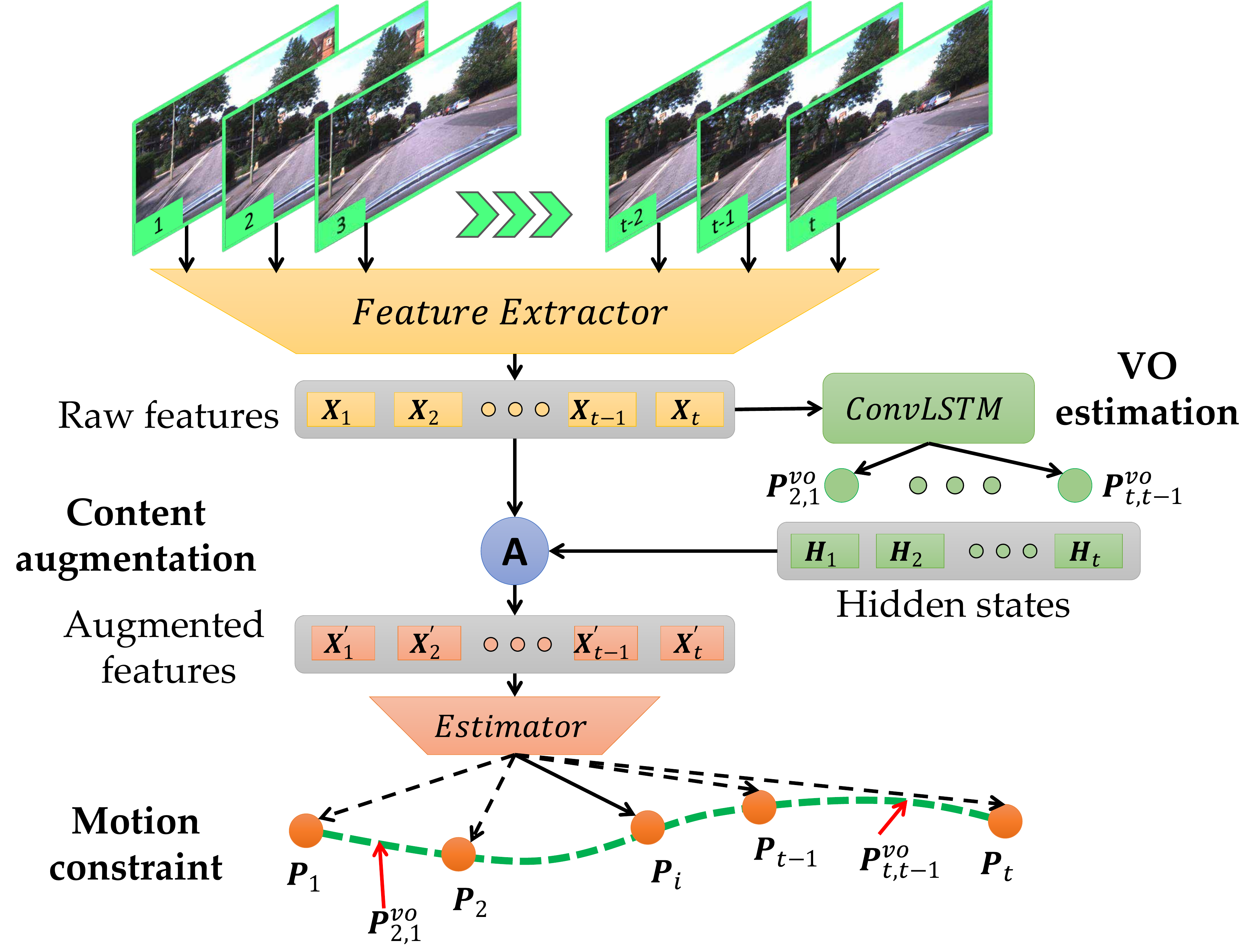}
	\caption{Overview of our framework. Sequential images are first encoded as high-level features which are then fed into the convolutional LSTM \cite{xingjian2015convolutional} for VO estimation. These features are augmented individually by searching co-visibility from hidden states in recurrent units before they are fed into the estimator for global pose regression. Predicted poses are optimized within a pose graph in which the predicted global poses and relative poses represent the nodes and edges, respectively.} 
	\label{fig:framework}
\end{figure}

Visual relocalization is a fundamental task in computer vision. With various applications in robotics, autonomous driving, and virtual/augmented reality, visual relocalization has been studied for decades and a considerable amount of geometry-based systems have been developed \cite{Taira_2018_CVPR, Toft_2018_ECCV, liu2017efficient, sattler2017efficient, sattler2016large}. Recently, with the success of deep learning techniques in computer vision tasks, global poses can be predicted by exploiting Convolutional Neural Networks (CNNs)~\cite{brahmbhatt2018mapnet, kendall2016modelling, kendall2017geometric, melekhov2017hourglass} and Recurrent Neural Networks (RNNs)~\cite{ walch2017pose-lstm, clark2017vidloc} in an end-to-end fashion. Most of these methods recover camera poses from single images. However, it is not easy to estimate stable poses from single images due to insufficient information or visual ambiguities such as repetitive textures, similar appearances, and the accuracy may also degrade with noise and over-exposure. As a sequence provides locally consistent spatio-temporal cues over consecutive frames, they have potentials to overcome limitations that single images based relocalization suffers.

In order to best utilize the local consistency, we delve
into the nature of sequential data in terms of content and
motion. First, in contrast to single images, a sequence can build a local map aggregating contents from consecutive frames from different viewpoints \cite{schonberger2016colmap}. Thus, a sequence
provides wider field of view and can distinguish stably detected features. Ambiguities arisen from single images can be alleviated with the local map. Additionally, considering the continuous nature of camera motions, relative poses can be exploited as additional constraints to mitigate the uncertainty that global pose estimation undergoes \cite{mur2017orb-slam2, engel2018dso}.

In this paper, we propose to learn camera relocalization from sequences by incorporating a deep VO component.
The VO sub-network accepts sequential images as input and predicts relative poses between consecutive frames. The hidden states that preserve historical information of sequence are taken as local maps. Specifically, our model learns camera relocalization by employing a two-step strategy. Raw features are first augmented according to the co-visibility in the local map for global pose prediction, namely content-augmentation module; VO estimation is then taken as a local motion constraint to enforce the consistency of predicted global camera poses within a pose graph.

The overview of our architecture is shown in~Fig.~\ref{fig:framework}. The encoder extracts features from images. The VO sub-network takes sequential features as input and calculates relative poses using convolutional LSTMs (Long Short-Term Memory) \cite{xingjian2015convolutional}. The extracted features are then augmented using hidden states with a soft attention, and fed into a pose predictor to regress camera poses for each view respectively. Finally, with vertexes denoting global poses and edges representing relative poses, a pose graph is established to enforce the consistency of predicted results. Our model learns camera relocalization and VO estimation jointly in a unified model, and can be used in an end-to-end fashion. Our contributions can be summarized as follows:


\begin{itemize}
	\item We propose to take advantages of the local information in a sequence to support global camera poses estimation by incorporating a VO component.
	\item Instead of raw features, our model estimates global poses from contents augmented based on covisibility in local maps.
	\item We adopt VO results as additional motion constraints to refine global poses within a pose graph during training and testing processes.
\end{itemize}

Our approach outperforms state-of-the-art methods in both the indoor 7-Scenes and outdoor Oxford RobotCar datasets, especially in challenging scenarios. The rest of this paper is organized as follows. In Sec.~\ref{related_work}, related works on relocalization and visual odometry are introduced. In Sec.~\ref{method}, our framework is described in detail. The performance of our approach is compared with previous methods in Sec.~\ref{experiment}. Finally, we conclude the paper in Sec.~\ref{conclusion}.

\section{Related work}
\label{related_work}
We first introduce relevant methods for camera relocalization, and then give a brief discussion on related VO algorithms.

\textbf{Visual Relocalization} Image-based relocalization algorithms recover the global pose approximately using the one of the most similar image in the database \cite{jegou2010vald, Balntas_2018_ECCV}. The approximation-like nature results in low accuracy. Structure-based algorithms require a scene 3D representation and search correspondences between 3D points and extracted features from a query image to establish 2D-3D matches \cite{Taira_2018_CVPR, Brachmann_2018_CVPR, brachmann2017dsac}. Then, the camera poses are estimated by applying RANSAC and solving a Perspective-n-Point problem \cite{camposeco2017toroidal, sattler2017efficient, kendall2017geometric}. Despite their promising performance, they require high computational cost, accurate intrinsic calibration and initialization for projection.


Recent works show that global camera poses can be estimated using a neural network in an end-to-end fashion. PoseNet \cite{kendall2015posenet} is the first to regress 6-DoF camera poses directly with deep neural networks in an end-to-end fashion. A series of following works extend PoseNet by modeling the uncertainty of poses with Bayesian CNNs \cite{kendall2016modelling}, introducing the geometric loss \cite{kendall2017geometric}, and correcting the structured features using LSTMs \cite{walch2017pose-lstm}. Melekhov \etal \cite{melekhov2017hourglass} add skip connections on ResNet34 \cite{he2016resnet} for preserving the fine-grained information. All abovementioned methods estimate camera localization from single images.

VidLoc \cite{clark2017vidloc} and MapNet \cite{brahmbhatt2018mapnet} are closely related to our work. VidLoc \cite{clark2017vidloc} accepts video clips as input and adopts regular bidirectional LSTMs to model the sequence. Although LSTMs can partially enhance observations, it cannot remember historical knowledge for a long time \cite{sukhbaatar2015memory}, resulting in poor performance in processing long sequences. We instead employ a soft attention mechanism to remedy the finite capacity of recurrent units. Moreover, additional
motion constraints are exploited to enforce the consistency of poses, which is ignored in VidLoc \cite{clark2017vidloc}. MapNet \cite{brahmbhatt2018mapnet} enforces the motion consistency between predicted global poses during training. However, the input for each view is only a single image, and it requires extra data obtained from GPS or VO/SLAM systems \cite{mur2017orb-slam2, engel2018dso} for pose refinement. In contrast, our model takes both the contents and motion cues from a sequence into consideration. Besides, the motion constraints can be performed in both training and testing stages, as our model directly provides relative poses from the VO component.

\textbf{Visual Odometry} Traditionally, visual odometry is solved by minimizing reprojection \cite{mur2017orb-slam2} or photometric errors \cite{engel2018dso, engel2014lsd-slam}. Recently, a number of learning-based VO systems have been developed. Some leverage CNNs to predict ego-motions from videos \cite{yin2018geonet, zhou2017egomotion} or relative poses between paired images \cite{zhou2018deeptam, ummenhofer2017demon}. VO can also be formulated as sequence-to-sequence problem and solved by a LSTM for sequence modeling \cite{wang2017deepvo, xue2018fea, xue2019beyond}. Benefited from the local temporal information preserved in LSTMs, relative results are predicted with promising accuracy. 

In this paper, we embed a deep VO component to assist relocalization. The hidden states with temporal information are taken as local maps to augment the raw features of each view. Moreover, the relative poses obtained from the VO component are exploited as motion constraints to enforce the consistency of global poses within a pose graph.



\section{Method}
\label{method}

In this section, we introduce our deep camera relocalization framework in detail. An overall network is illustrated
in Fig.~\ref{fig:network}. The feature extractor module encodes images into high-level features for VO estimation in Sec.~\ref{mtd:feature}. The extracted contents are augmented for global poses prediction in Sec.~\ref{mtd:content}. Estimated global poses are finally optimized
along with VO results within a pose graph in Sec.~\ref{mtd:motion}. 

\subsection{Feature Extraction and VO Estimation}
\label{mtd:feature}
\begin{figure}[t]
	\begin{center}
		\includegraphics[width=1.\linewidth]{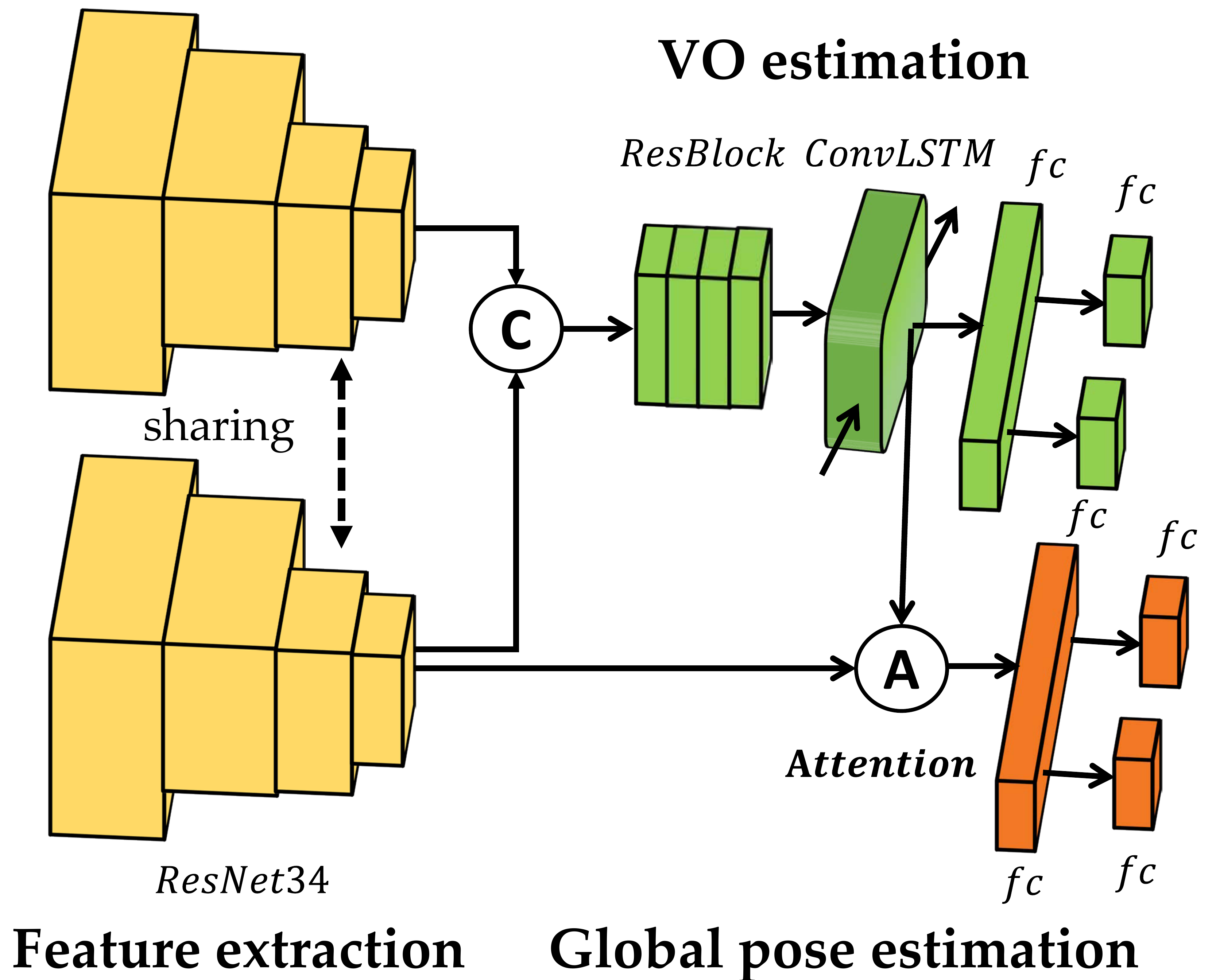}
	\end{center}
	\caption{ResNet34 \cite{he2016resnet} with the last two layers discarded is used to extract features with 512 channels.  Features of two consecutive frames are concatenated and passed through the \textit{ResBlock} consisting of 3 \textit{BasicBlock}s for fusion. Fused features are then fed into a convolutional LSTM \cite{xingjian2015convolutional} for VO estimation. Raw features are augmented based on the co-visibility in local maps using a soft attention mechanism. Outputs of recurrent units and augmented features are vectorized to calculate the relative and global 6-DoF poses via 3 FC layers, respectively. The output channels of the first FC layer for both global and relative poses estimation are 2048, while the left are 3.
	} 
	\label{fig:network}
\end{figure}

\textbf{Feature Extraction} Similar to \cite{brahmbhatt2018mapnet, melekhov2017hourglass}, we adopt the modified ResNet34 \cite{he2016resnet} to extract features from images. The last global average pooling (GAP) layer and fully-connected (FC) layer are discarded, producing 3D-tensors as feature maps. As shown in Fig.~\ref{fig:network}, since camera relocalization and visual odometry are two closely related sub-tasks, the similarity can also be inherited in the feature space by sharing weights of feature extraction module. To be specific, the output of our modified ResNet34 for frame $t$ is taken as raw feature $X_t$. Two consecutive raw features are concatenated along the channel dimension and passed through a group of residual networks to obtain fused features $X^{vo}_t$ for the VO task. This group of residual networks shares the same number of \textit{BasicBlock} with the last residual layer in ResNet34. It is worthy to note that the number of input channels is changed and the pooling operation is disabled. 

\textbf{Visual Odometry Estimation} Learning visual odometry from image sequences using LSTMs \cite{hochreiter1997lstm} for temporal reasoning has been proved effective \cite{wang2017deepvo, xue2018fea}. The regular LSTM \cite{hochreiter1997lstm} cannot retain the spatial connections of features with only 1D vectors as input. We adopt the convolutional LSTM \cite{xingjian2015convolutional} as our recurrent unit following the work in \cite{xue2018fea}. The fused feature $X^{vo}_t$ is fed into the recurrent unit as:
\ 
\begin{align}
	H_t, O_t = \mathcal{U}(X^{vo}_t, H_{t-1}) \ ,
\end{align}
where $H_{t-1}$ and $H_t$ represent hidden states at time $t - 1$ and $t$, respectively. $O_t$ denotes the output 3D tensor of the recurrent unit $\mathcal{U}$ at time $t$. It is passed through a GAP layer and three FC layers (with output channels of 2048, 3 and 3) to calculate the 6-DoF relative pose $P^{vo}_{t,t-1}$. 

Intuitively, the feature flow passing through recurrent units is filtered and fused via the \textit{gates} in convolutional LSTMs. Hidden states with historical knowledge preserved are viewed as local maps to boost the relocalization task. 


\subsection{Content-augmented Pose Estimation}
\label{mtd:content}
We aim at utilizing multiple observations from a sequence
to reduce the ambiguity of a single image. Since images are high-dimension data with much redundant information, learning information from raw data is ineffective \cite{wang2017deepvo}. On the other hand, LSTMs cannot preserve long-term knowledge \cite{sukhbaatar2015memory}. Hence, fusing features instead of images using bidirectional LSTMs is incapable of processing long sequences. To deal with the problem, we take inspirations from classic SLAM systems \cite{mur2017orb-slam2} by searching the co-visibility relations from local maps built in the VO stream.

\begin{figure}[t]
	\begin{center}
		\includegraphics[width=1.\linewidth]{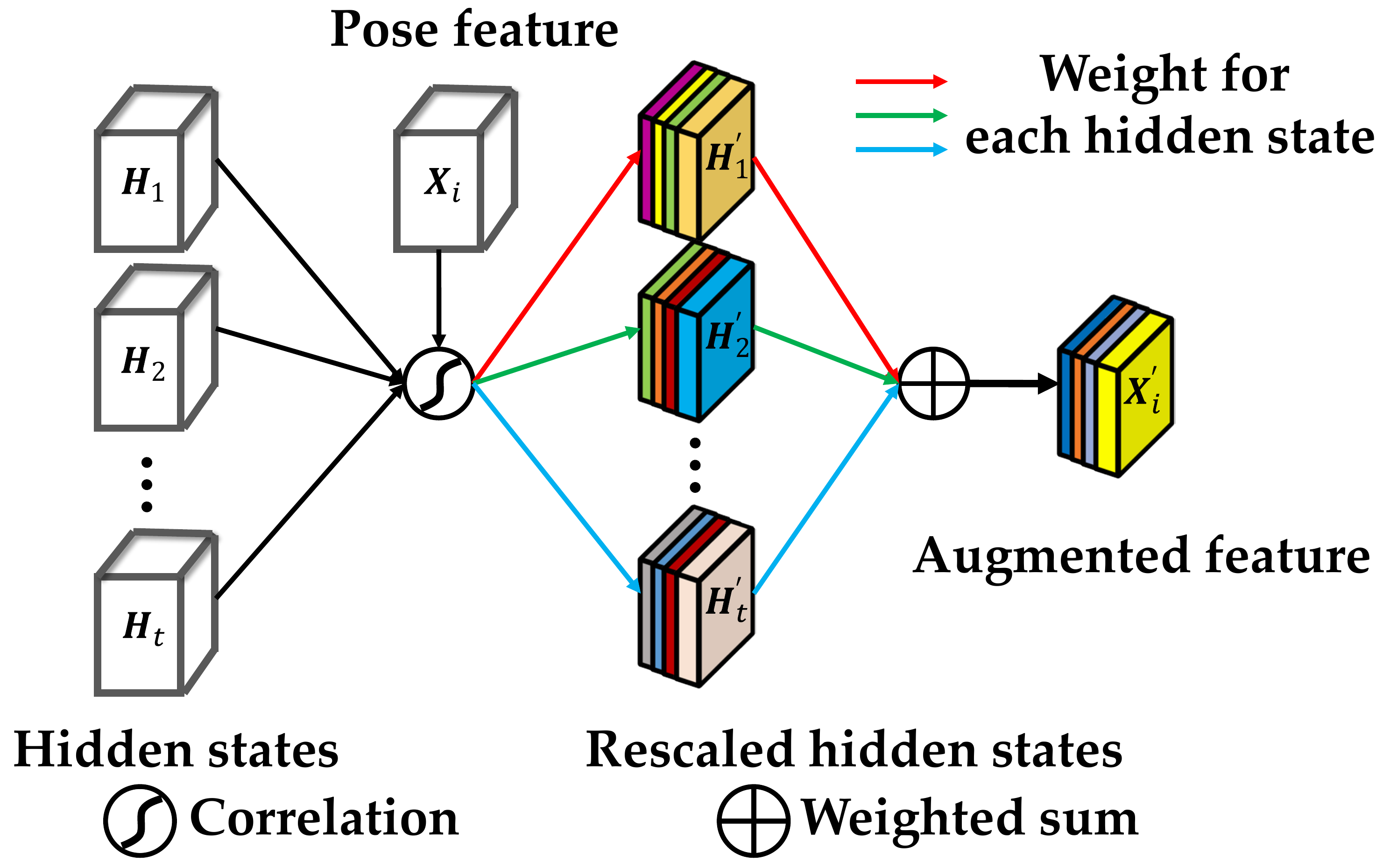}
	\end{center}
	\caption{We search the co-visible contents from all hidden states with the guidance of raw features. The selection is a soft attention by re-scaling each hidden state in temporal domain and each channel of hidden states in spatial domain, respectively.} 
	\label{fig:content}
\end{figure}

\textbf{Content Augmentation} In our work, both observations and local maps are represented implicitly as features. Hence, the correlation in the feature space represents the co-visibility relations between local maps and extracted features for each view. We perform the correlation
by employing a soft attention mechanism with raw features as guidance. Considering the fact that the capacity
of recurrent units is finite, an attention module is operated
in both spatial and temporal domains, as shown in
Fig.~\ref{fig:content}.


More specifically, for each observation $I_t$, we take its raw feature map $X_t$ as the guidance to select the most related knowledge by re-weighting all hidden states in the spatio-temporal domain. Since both $X_t$ and $H_{i}$ are 3D tensors, following GFS-VO \cite{xue2018fea}, we calculate the cosine similarity between the vectorized feature map of $X_t$ and $H_i$ in each channel. Unlike GFS-VO that focuses on spatial correlation between two tensors, we extend the correlation to the temporal domain based on the intuition that each hidden state contributes discriminatively to different views. The whole process is described as:
\ 
\begin{align}
	X^{'}_t = \sum_{i=1}^{N}(\mathcal{A}^{T}(X_t, H_i)\sum_{j=1}^{C}\mathcal{A}^{S}(X^{j}_{t}, H^{j}_{i})H^{j}_{i}) \ .
\end{align}
Here, $\mathcal{A}^{T}$ and $\mathcal{A}^{S}$ denote the temporal and spatial correlation respectively. $N$ and $C$ are the number of hidden states and channels each hidden state has. $H^{j}_{i}$ denotes the $j$th channel of the $i$th hidden state. $X^{'}_t \in \mathbb{R}^{H \times W \times C}$ is the obtained input for $I_t$ by fusing co-visible contents from the whole sequence explicitly in the feature space.

The soft attention we adopt discovers the filtered co-visible contents and retains observations beyond a single view but appear in local maps. More importantly, the obtained content for each frame is view related and the field of view is enlarged implicitly by images in the whole sequence in feature space. 

\textbf{Global Pose Estimation} Given the augmented feature $X^{'}_t$ at time $t$, we can estimate the corresponding global pose with the \textit{Estimator}.  $X^{'}_t$ is fused by two convolutional layers with kernel size of 3. Then the feature is fed into a GAP layer and three FC layers (with output channels of 2048, 3 and 3), mapping features to global poses (see Fig.~\ref{fig:network}). As FC layers preserve much of the global map information \cite{brahmbhatt2018mapnet}, the process can be viewed as a registration between the \textit{local map} from a specific view and the \textit{global map} of the scene. 

\subsection{Motion-based Refinement}
\label{mtd:motion}
\begin{figure}[t]
	\begin{center}
		\includegraphics[width=1.\linewidth]{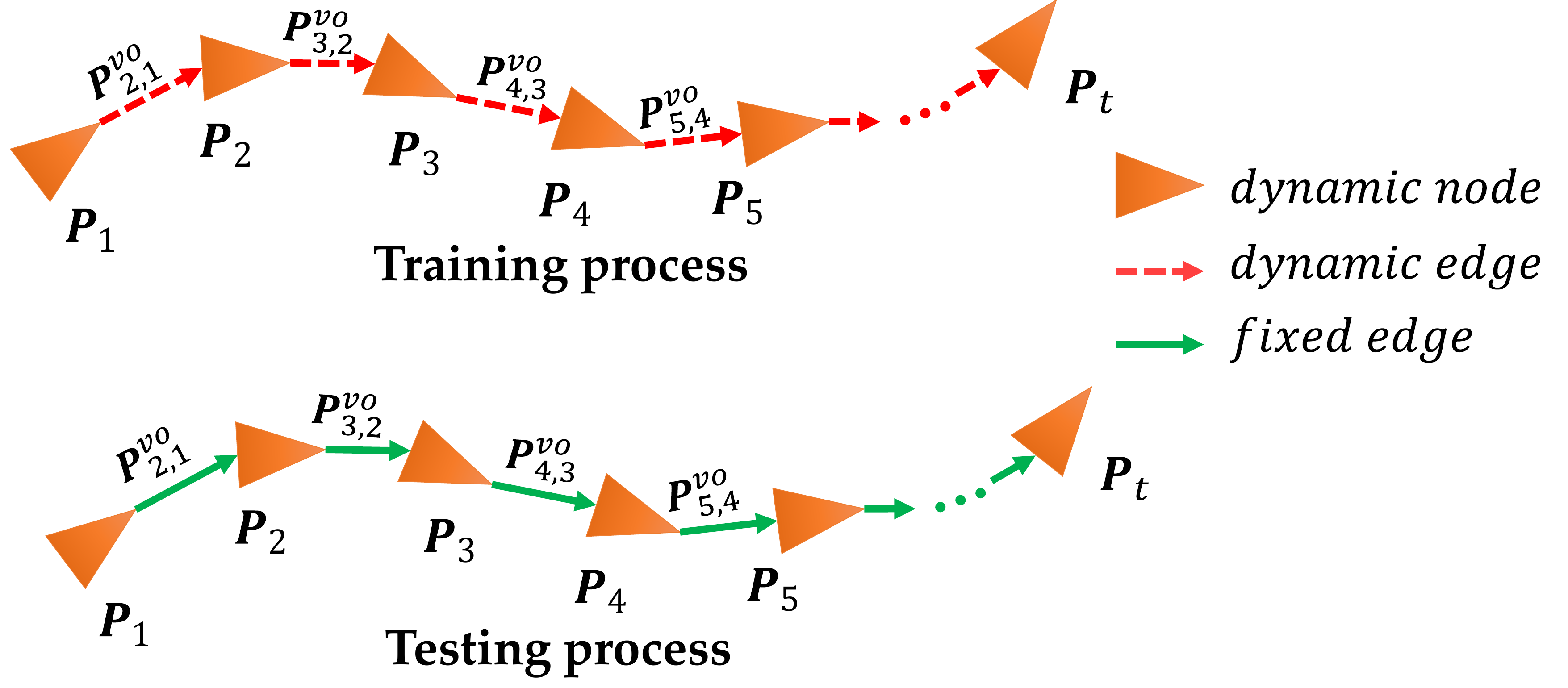}
	\end{center}
	\caption{The pose graph consists of global poses indicating nodes and VO results representing edges. During training (top), both the nodes and edges are optimized and updated dynamically. While in testing (bottom), we perform a standard pose graph optimization \cite{calafiore2016pose-graph} to optimize only the global poses with relative poses fixed as marginalization.} 
	\label{fig:motion}
\end{figure}

The VO estimation provides relative poses with promising accuracy but drifts over time. The predicted global poses undergo uncertainties while are drift-free. Therefore, the two sub-tasks can be learned cooperatively, resulting in a win-win situation. In our work, we employ a pose graph to enforce the consistency of predictions. In the graph, all global poses are defined as nodes which are connected by edges, representing the relative poses, as shown in Fig.~\ref{fig:motion}.

\textbf{Joint Optimization in Training}
During the training process, since the global and relative poses are not optimal, both nodes and edges are optimized and updated dynamically. We consider a local window with all frames in the videos, and design the joint loss with the global and local poses coupled as:
\ 
\begin{align}
	\mathcal{L}_{joint} &= \sum_{i=1}^{N-1}\mathcal{D}(P_{i+1}, P^{vo}_{i+1,i}P_{i}) \ .
\end{align}
The $\mathcal{D}$ computes the distance between two poses (to be discussed further). $\mathcal{L}_{joint}$ indicates the consistency between predicted global and relative poses. As the joint loss mainly enforces the consistency between predicted poses, it is combined with additional losses to optimize the model in a supervised manner.

\textbf{Pose Graph Optimization in Testing}
When in testing, the VO sub-network recovers locally accurate relative poses between consecutive frames. To further reduce the uncertainty of global poses, we perform a standard pose graph optimization (PGO) \cite{calafiore2016pose-graph} by optimizing only the nodes with edges fixed. We minimize the error as:
\ 
\begin{align}
	\min_{P^{'}_{1:N}} \sum_{i=1}^{N}\mathcal{D}(P_i, P^{'}_i) + \alpha\sum_{i=1}^{N-1}\mathcal{D}(P^{'}_{i+1}, P^{vo}_{i+1,i}P^{'}_{i}) \ .
\end{align}

$P_i$ and $P^{vo}_{i+1,i}$ are the predicted global and local poses. $P^{'}_i$ denotes global poses to be optimized. As no loop closures are provided as in classic SLAM systems \cite{mur2017orb-slam2}, the predicted global poses are optimized with local poses as constraints to force the  consistency. The error against original predictions is added, restricting the final results from deviating too much from the original predictions. $\alpha$ is used to balance the two errors.

Though MapNet \cite{brahmbhatt2018mapnet} also employs a pose graph optimization during inference, it requires extra computation with known relative poses of testing images from the ground truth or other VO/SLAM systems \cite{engel2018dso}. In contrast, our approach obtains these results from the model, and thus gets rid of the dependence on extra data.


\subsection{Loss Function} 
As our network learns camera relocalization and VO estimation jointly in a unified model, the losses are defined on both the global and relative poses as:
\ 
\begin{align}
	\mathcal{L}_{g} &= \sum_{i=1}^{N} \mathcal{D}(\hat{P}_i, P_i) \ , \\
	\mathcal{L}_{vo} &= \sum_{i=2}^{N} \mathcal{D}(\hat{P}^{vo}_{i+1,i}, P^{vo}_{i+1,i}) \ .
\end{align} 
$\mathcal{L}_{g}$ and  $\mathcal{L}_{vo}$ represent the global and relative losses, respectively. $\hat{P}_i$ and  $\hat{P}^{vo}_{i}$ are ground truth global and relative poses. For function $\mathcal{D}$, we adopt the definition in \cite{kendall2017geometric}:
\ 
\begin{align}
	\mathcal{D}(\hat{P}, P) &= ||\hat{\mathbf{t}} - \mathbf{t}||_1\exp^{-\beta} + \beta + ||\hat{\mathbf{w}} - \mathbf{w}||_1\exp^{-\gamma} + \gamma \ ,
\end{align}
where $\mathbf{t}$ and $\mathbf{w}$ are both 3-DoF parameters representing the predicted translation and rotation (in formulation of log quaternion), while $\hat{\mathbf{t}}$ and $\hat{\mathbf{w}}$ denote the ground truths. $\beta$ and $\gamma$ are the weights to balance the translational and rotational errors. During training, $\beta$ and $\gamma$ are not fixed, we optimize them as parameters with an initialization $\beta^0$ and $\gamma^0$.

The total loss consists of global pose loss $\mathcal{L}_{g}$, VO loss $\mathcal{L}_{vo}$ and motion constraint loss $\mathcal{L}_{joint}$ as:
\ 
\begin{align}
	\mathcal{L}_{total} &= \mathcal{L}_{g} + \mathcal{L}_{vo} +\mathcal{L}_{joint} \ .
\end{align}

Although different losses have different scales, the parameters $\beta$ and $\gamma$ of each loss can balance the scale.


\section{Experiments}
\label{experiment}
\setlength{\tabcolsep}{4.pt}
\begin{table*}[t]
	\footnotesize
	\centering
	\setlength{\abovecaptionskip}{0pt}%
	\setlength{\belowcaptionskip}{0pt}%
		\begin{center}
			\begin{tabular}{lcccccccc}
				\hline
				\hline
				& \multicolumn{8}{c}{Sequence} \\
				Method & Chess & Fire & Heads & Office & Pumpkin & Kitchen & Stairs & Avg \\
				\hline
				
				PoseNet15 \cite{kendall2015posenet} & 0.32
				m, $8.12^\circ$ & 0.47m, $14.4^\circ$ & 0.29m, $12.0^\circ$ & 0.48m, $7.68^\circ$ & 0.47m, $8.42^\circ$ & 0.59m, $8.64^\circ$ & 0.47m, $13.8^\circ$ & 0.44m, $10.4^\circ$ \\
				PoseNet16 \cite{kendall2016modelling} & 0.37
				m, $7.24^\circ$ & 0.43m, $13.7^\circ$ & 0.31m, $12.0^\circ$ & 0.48m, $8.04^\circ$ & 0.61m, $7.08^\circ$ & 0.58m, $7.54^\circ$ & 0.48m, $13.1^\circ$ & 0.47m, $9.81^\circ$ \\
				PoseNet17 \cite{kendall2017geometric} & 0.13m, $4.48^\circ$ & 0.27m, $11.30^\circ$ & 0.17m, $13.00^\circ$ & 0.19m, $5.55^\circ$ & 0.26m, $4.75^\circ$ & 0.23m, $5.35^\circ$ & 0.35m, $12.40^\circ$ & 0.23m, $8.12^\circ$ \\
				Hourglass \cite{melekhov2017hourglass} & 0.15m, $6.17^\circ$ & 0.27m, $\mathbf{10.84}^\circ$ & 0.19m, $\mathbf{11.63}^\circ$ & 0.21m, $8.48^\circ$ & 0.25m, $7.01^\circ$ & 0.27m, $10.15^\circ$ & 0.29m, $12.46^\circ$ & 0.23m, $9.53^\circ$ \\
				LSTM-Pose \cite{walch2017pose-lstm} & 0.24m, $5.77^\circ$ & 0.34m, $11.9^\circ$ & 0.21m, $13.7^\circ$ & 0.30m, $8.08^\circ$ & 0.33m, $7.00^\circ$ & 0.37m, $8.83^\circ$ & 0.40m, $13.7^\circ$ & 0.31m, $9.85^\circ$ \\
				\textbf{Ours} & \textbf{0.09}m, $\mathbf{3.28}^\circ$ & \textbf{0.26}m, $10.92^\circ$ & \textbf{0.17}m, $12.70^\circ$ & \textbf{0.18}m, $\mathbf{5.45}^\circ$ & \textbf{0.20}m, $\mathbf{3.66}^\circ$ & \textbf{0.23}m, $\mathbf{4.92}^\circ$ & \textbf{0.23}m, $\mathbf{11.3}^\circ$ & \textbf{0.19}m, $\mathbf{7.47}^\circ$ \\
				\hline
				\hline
			\end{tabular}
		\end{center}%
	\caption{Median translation and rotation errors of previous methods using single images as input and our model on the 7-Scenes dataset \cite{shotton2013scene}. The best results are highlighted.}
	\label{tab:7scenes_image}
\end{table*}

\begin{figure*}[t]
	\def\subfig_width{0.23\textwidth}
\centering
	\begin{subfigure}{\subfig_width}
	\includegraphics[width=0.95\textwidth]{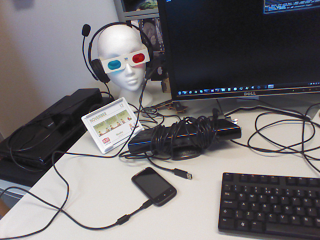}
\end{subfigure}%
\hfill
\begin{subfigure}{\subfig_width}
	\includegraphics[width=1\textwidth]{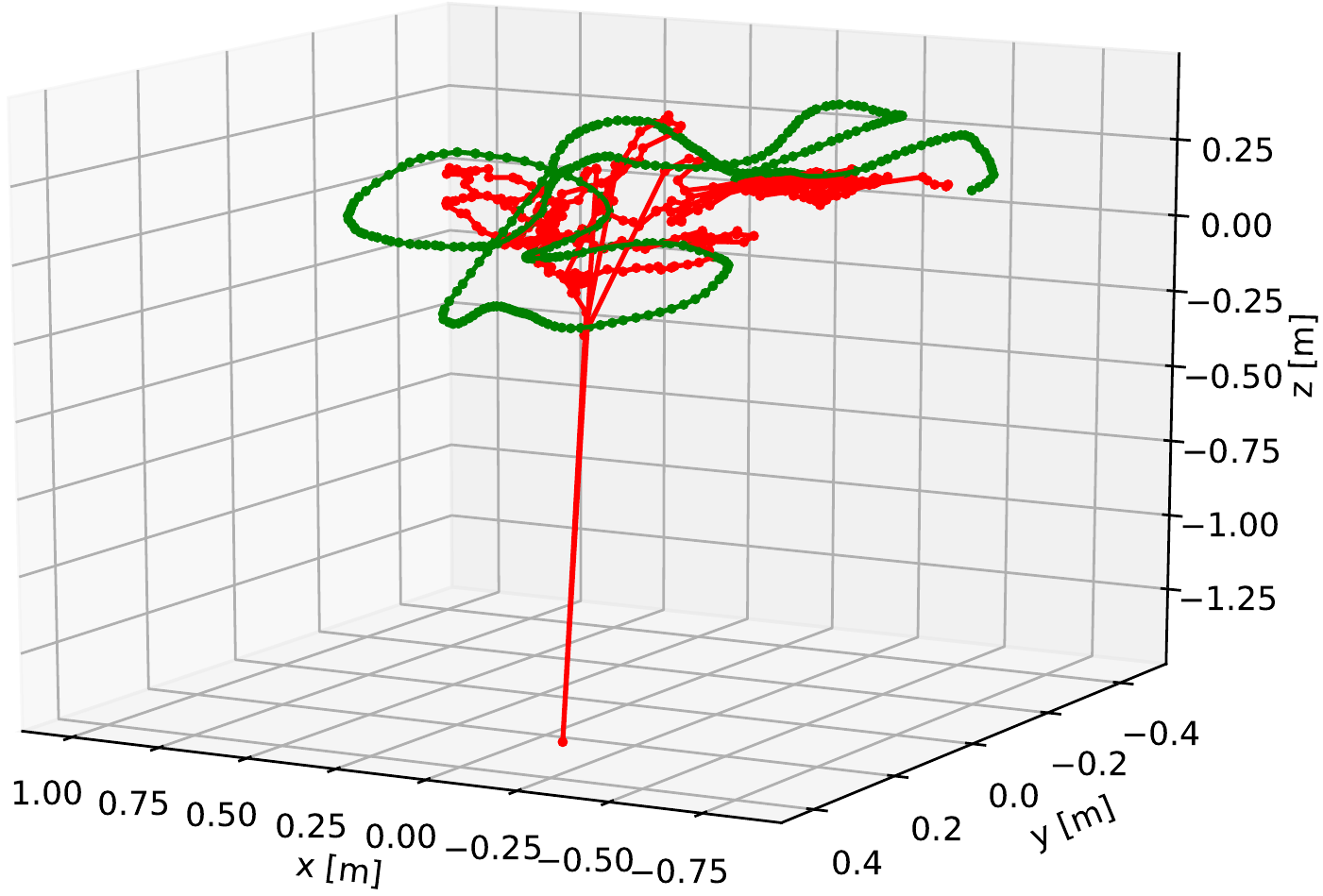}
\end{subfigure}%
\hfill
\begin{subfigure}{\subfig_width}
	\includegraphics[width=1\textwidth]{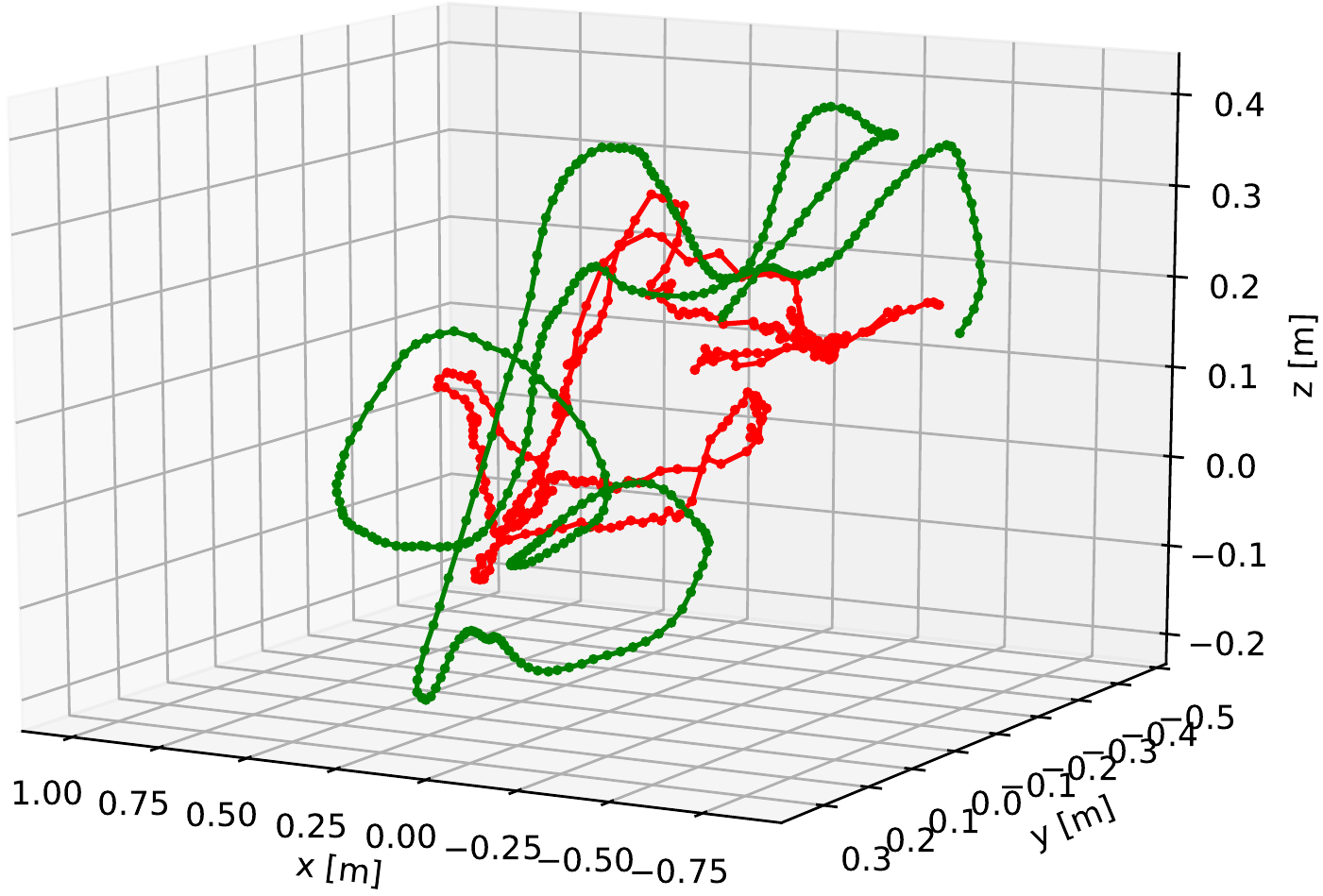}
\end{subfigure}%
\hfill
\begin{subfigure}{\subfig_width}
	\includegraphics[width=1\textwidth]{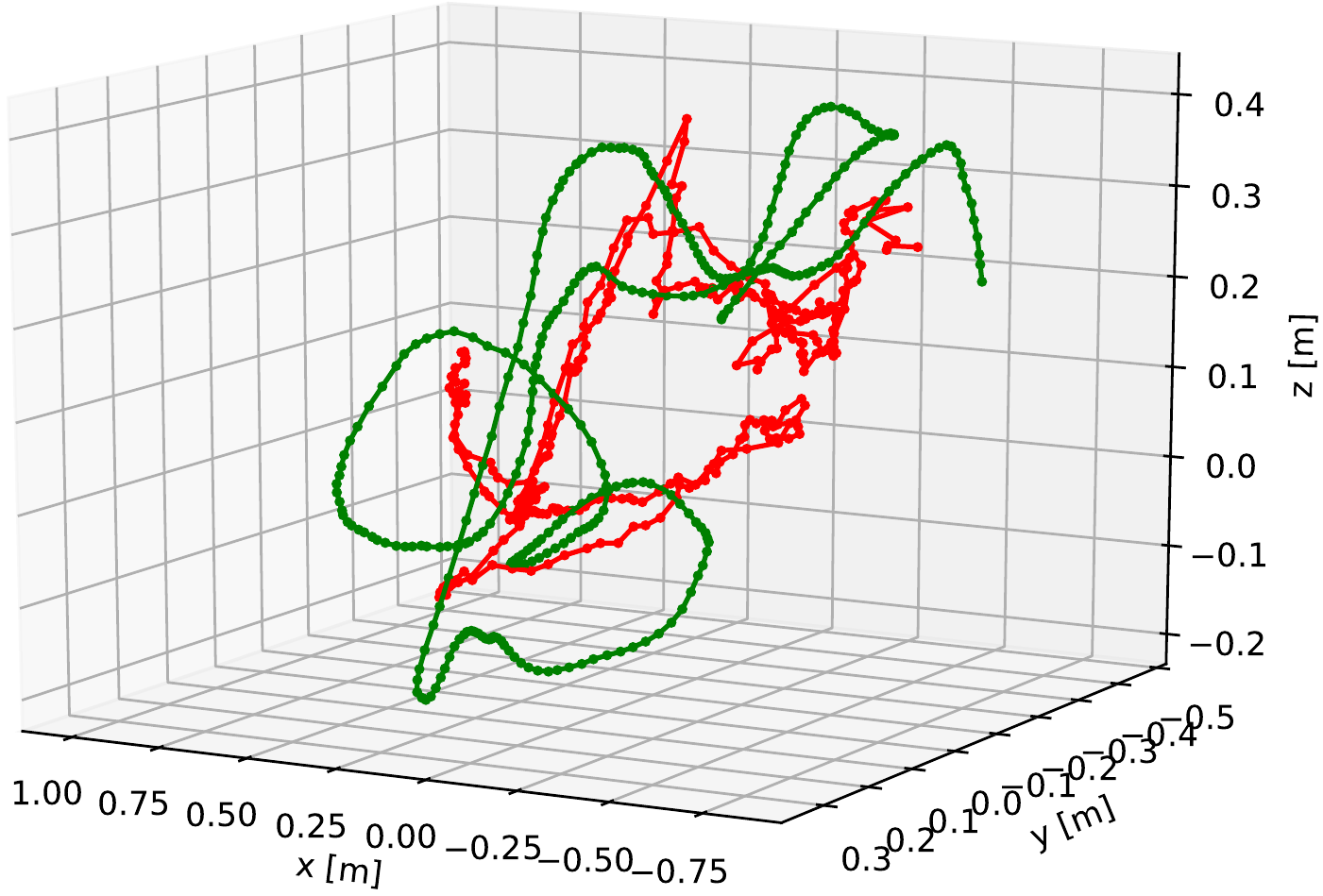}
\end{subfigure}%
\hfill

	\begin{subfigure}{\subfig_width}
	\includegraphics[width=0.95\textwidth]{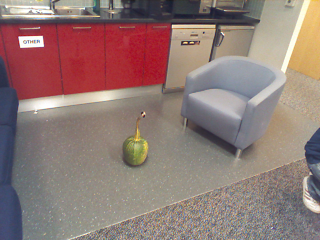}
\end{subfigure}%
\hfill
\begin{subfigure}{\subfig_width}
	\includegraphics[width=1\textwidth]{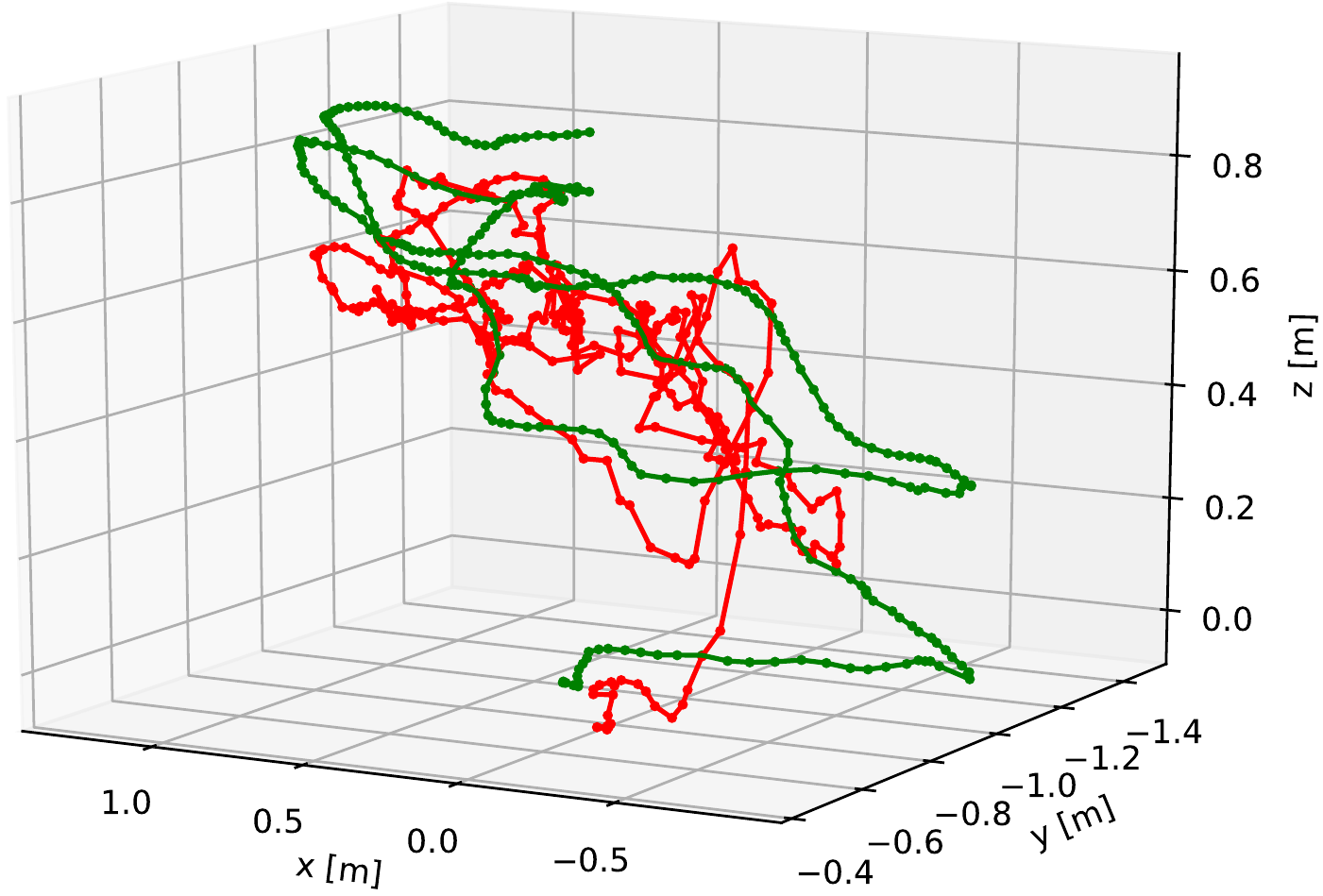}
\end{subfigure}%
\hfill
\begin{subfigure}{\subfig_width}
	\includegraphics[width=1\textwidth]{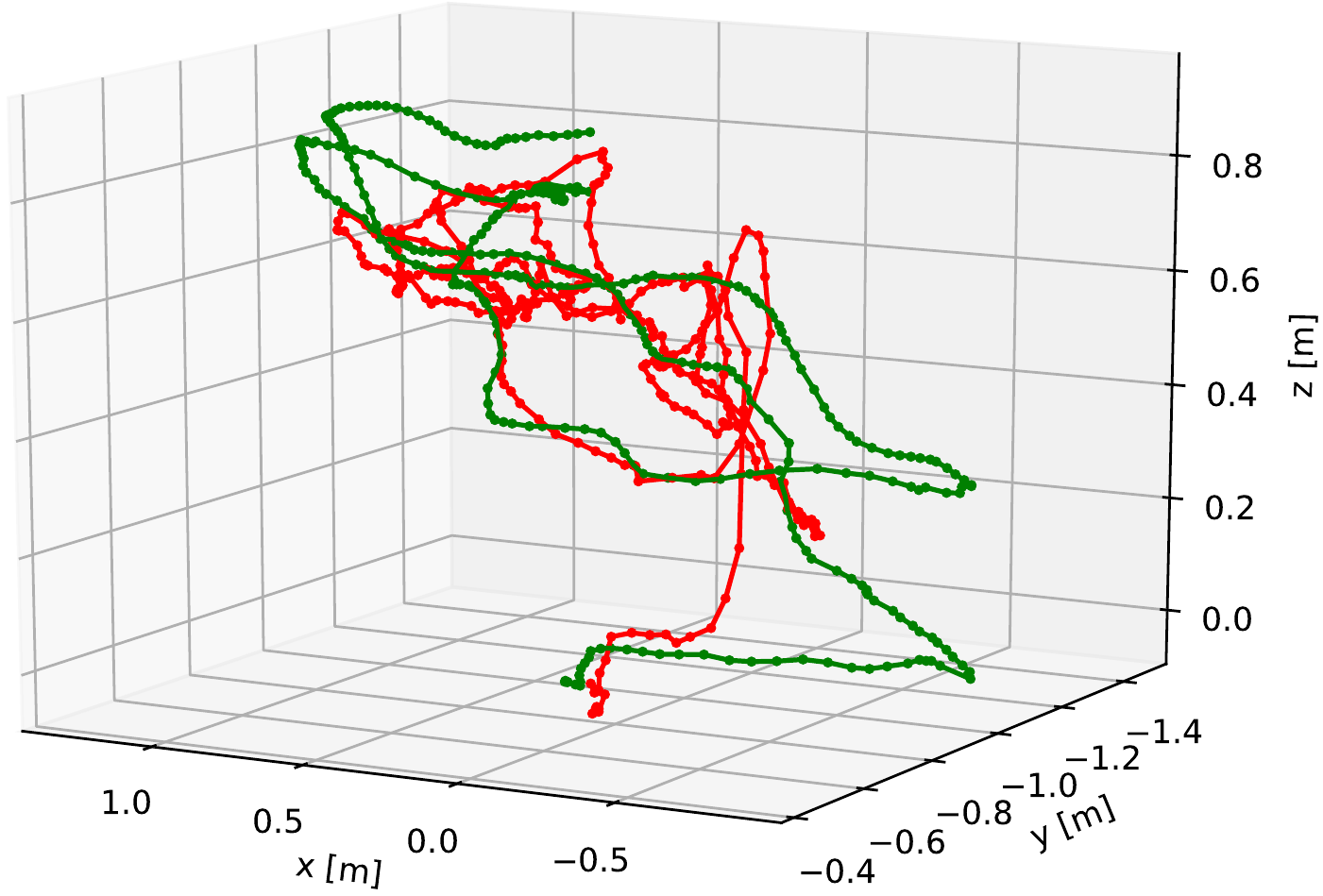}
\end{subfigure}%
\hfill
\begin{subfigure}{\subfig_width}
	\includegraphics[width=1\textwidth]{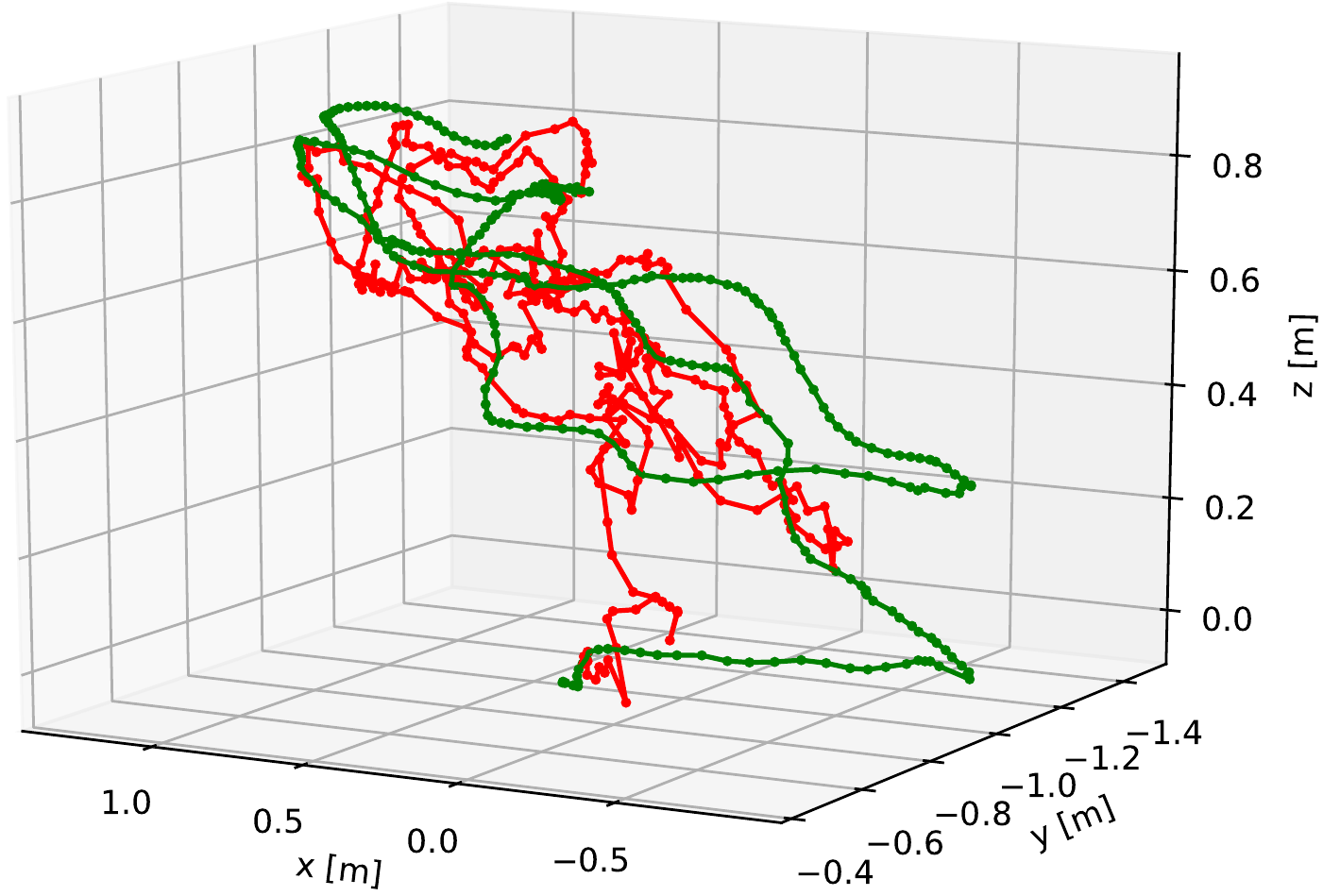}
\end{subfigure}

	\begin{subfigure}{\subfig_width}
	\includegraphics[width=0.95\textwidth]{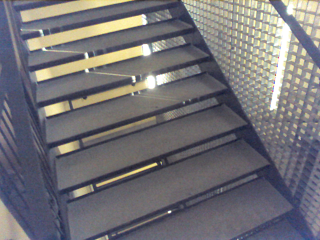}
	\centerline{Raw images}
\end{subfigure}%
\hfill
\begin{subfigure}{\subfig_width}
	\includegraphics[width=1\textwidth]{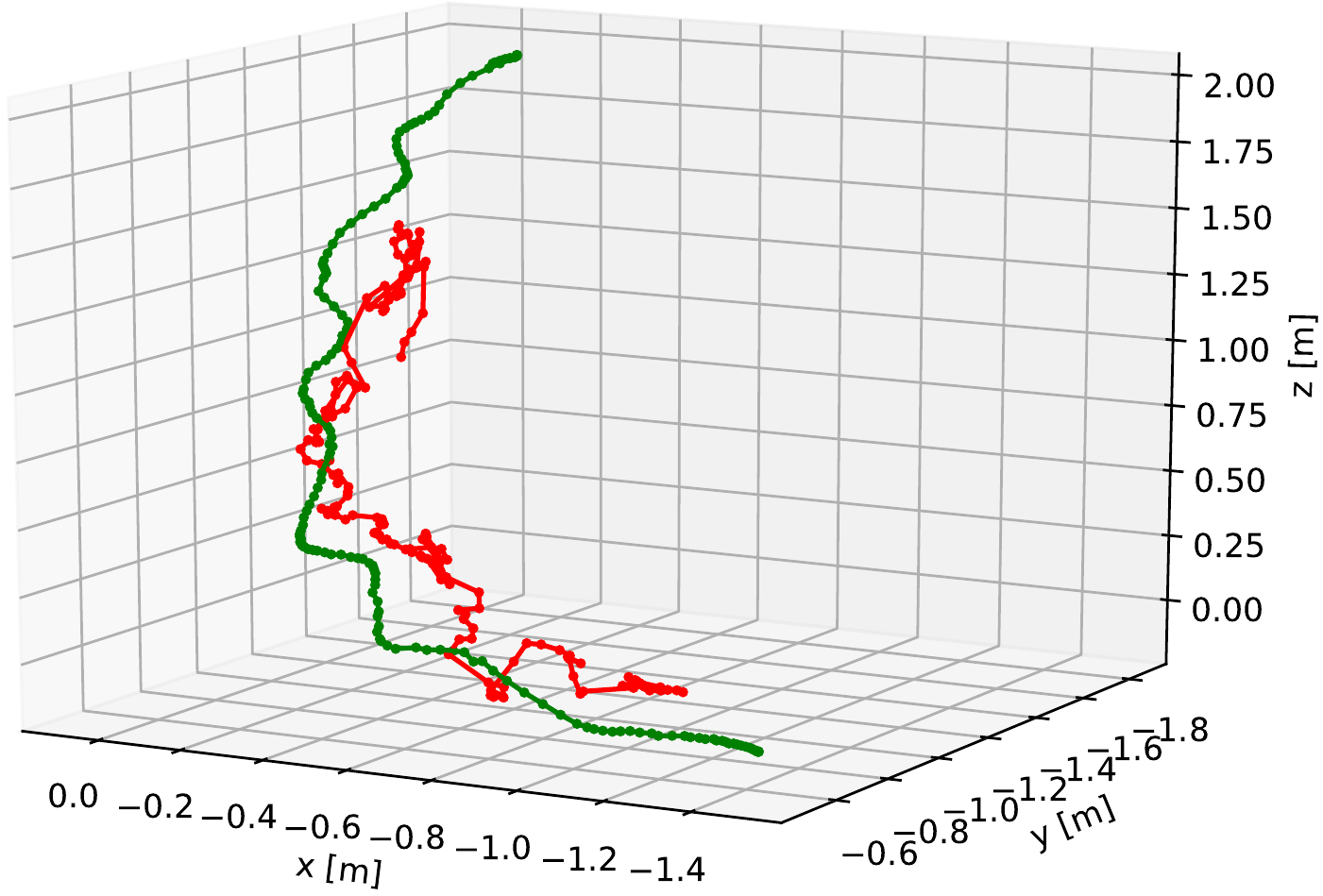}
	\centerline{PoseNet \cite{kendall2015posenet, kendall2016modelling, kendall2017geometric}}
\end{subfigure}%
\hfill
\begin{subfigure}{\subfig_width}
	\includegraphics[width=1\textwidth]{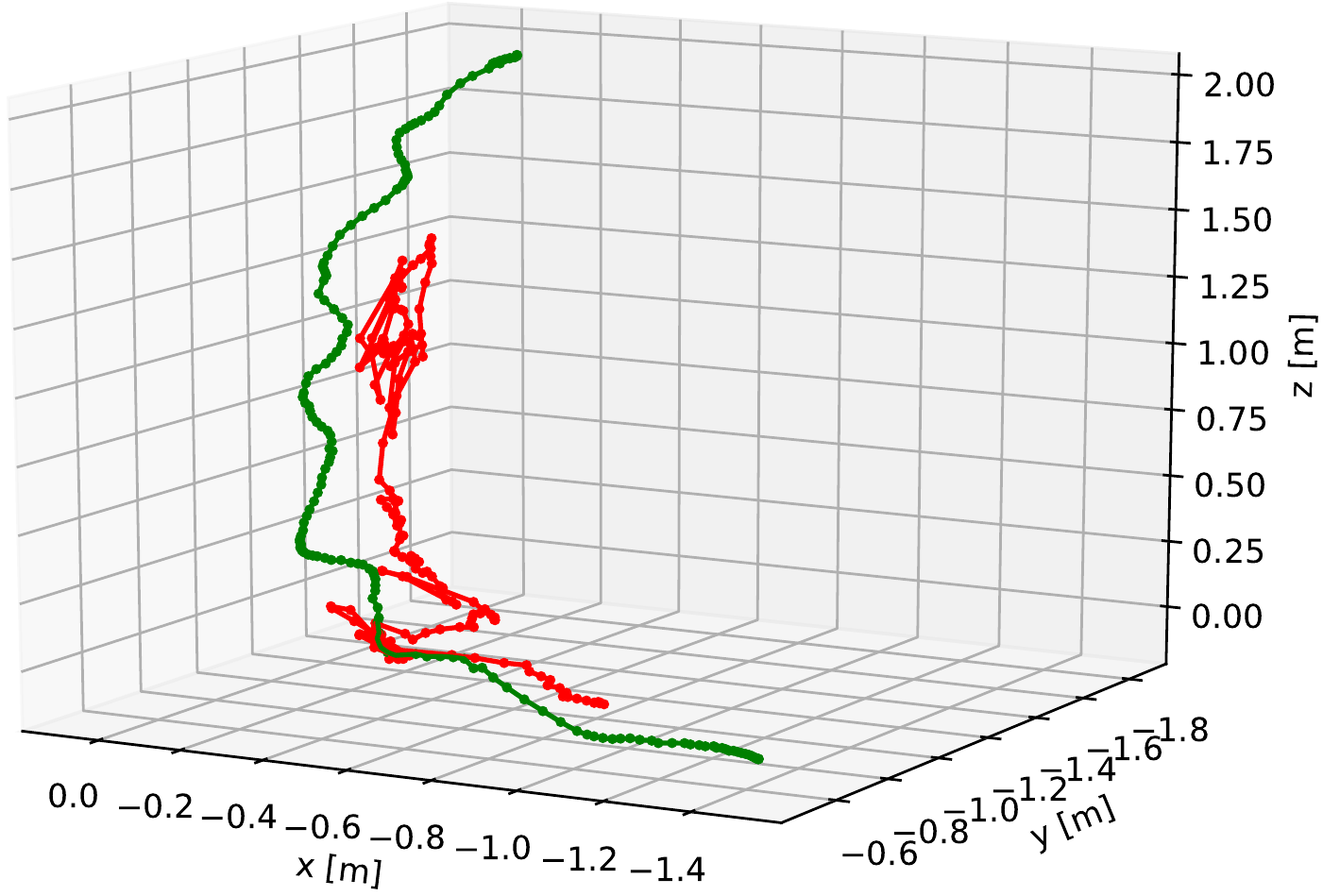}
	\centerline{MapNet \cite{brahmbhatt2018mapnet}}
\end{subfigure}%
\hfill
\begin{subfigure}{\subfig_width}
	\includegraphics[width=1\textwidth]{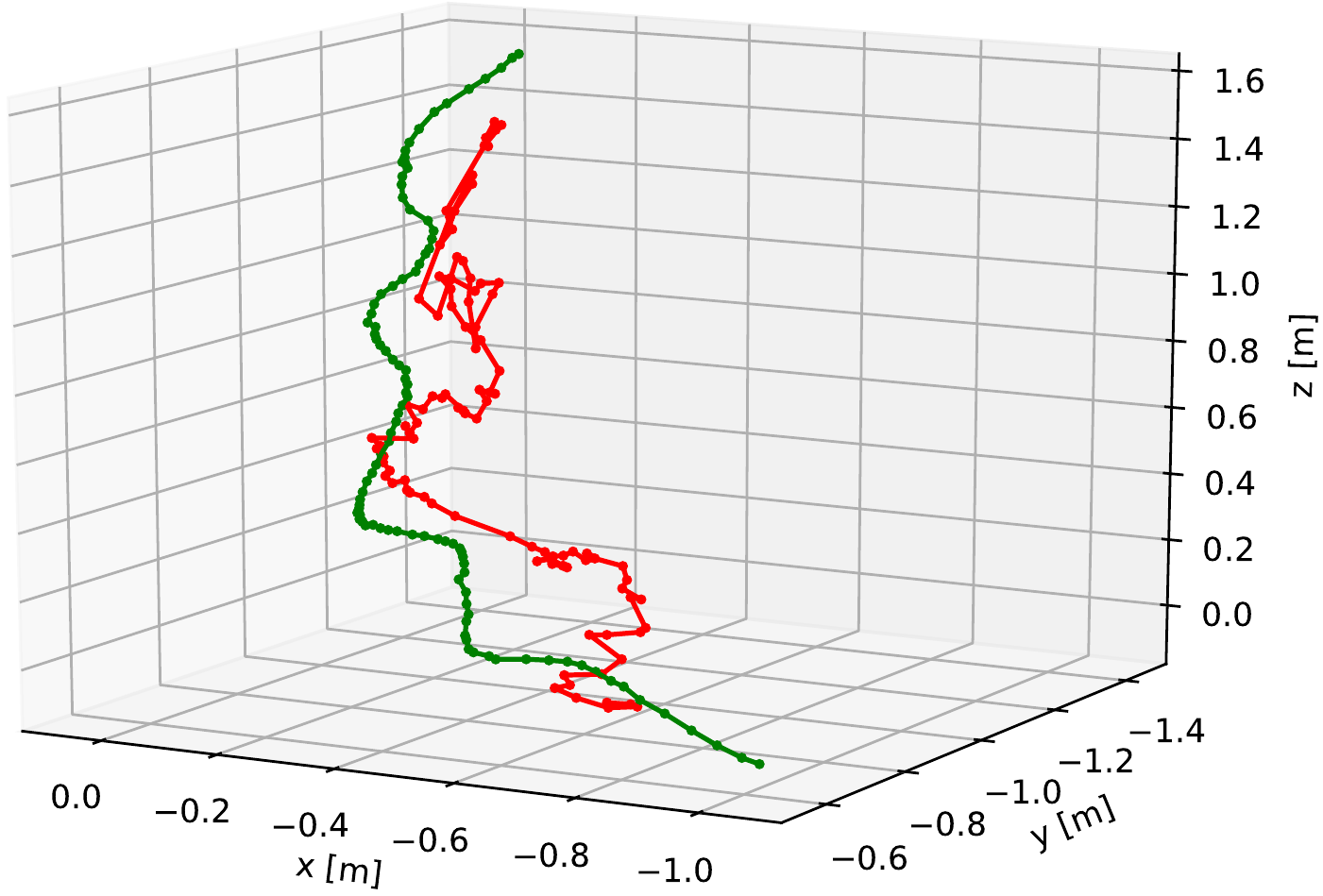}
	\centerline{Ours}
\end{subfigure}

\caption{Results on the 7-Scenes dataset \cite{shotton2013scene}. The green line indicates the ground truth and the red line represents the predicted trajectories of various mehtods. From top to bottom, three testing sequences are Heads-seq-01, Pumpkin-seq-07, Stairs-seq-01. The \textit{Pumpkin} contains many texture-less regions, and \textit{Stairs} has lots of highly repetitive textures.}
\label{fig:7scenes}
\end{figure*}
\setlength{\tabcolsep}{1.4pt}

In this section, we first introduce the implementation details, datasets used for evaluation and baseline methods for comparison. Then we compare our model with previous approaches in Sec.~\ref{exp:7scenes} and \ref{exp:robotcar}. The ablation study and additional results can be found in the supplementary material.



\textbf{Implementation Details}
\label{exp:implementation}
Our network takes monocular RGB image sequences as input.  We use 7 consecutive images to construct a sequence. Similar to \cite{brahmbhatt2018mapnet}, shorter side length of all input images are rescaled to 256. We calculate the pixel-wise mean for each of the scenes in the datasets and subtract them with the input images. All $\beta$s and $\gamma$s are set to -3 and 0,  respectively. The ResNet34 is pretrained on the ImageNet, while other parts of the network are initialized using MSRA method \cite{he2015msra}. We adopt the PyTorch \cite{pytorch} to implement the model on an NVIDIA 1080Ti GPU. Adam \cite{Kingma2014Adam} with weight decay of $5 \times 10^{-4}$  is used to optimize the network with batch size of 16 for 200 epochs in total. The initial learning rate is set to $10^{-4}$ and is kept during the whole training process.

\textbf{Dataset} We conduct extensive experiments on the popular 7-Scenes \cite{shotton2013scene} and Oxford RobotCar \cite{robotcar2017} datasets. The 7-Scenes was collected using a Kinect in seven different indoor environments with spatial extent less than 4 meters. Among the 7 categories, the \textit{Pumpkin} was recorded in the scene with a pumpkin on the floor, and contains many texture-less areas. The \textit{Stairs} was collected over the stairs with lots of highly repetitive textures. Both the \textit{Pumpkin} and \textit{Stairs} are challenging for relocalization algorithms (see Fig.~\ref{fig:7scenes}). Each category contains multiple sequences, some of which are used for training and the others for testing. The ground truth was obtained with KinectFusion. Both RGB and depth image sequences are provided, while only the RGB images are utilized in this paper.

The Oxford RobotCar dataset was captured through central Oxford over several periods in a year. Consequently, it contains observations under various conditions of weather, traffic, pedestrians, construction and roadworks, which makes it a big challenge for relocalization algorithms. Images captured by the center camera are used as input and the interpolations of INS measurements are used as the ground truth. We follow the train/split in MapNet \cite{brahmbhatt2018mapnet} by using two groups of subsets denoted as LOOP and FULL scenes.

\textbf{Baseline Methods} When evaluating on the 7-Scenes benchmark, the baseline algorithms include PoseNet15 \cite{kendall2015posenet}, PoseNet16 \cite{kendall2016modelling}, PoseNet17 \cite{kendall2017geometric}, Hourglass \cite{melekhov2017hourglass}, LSTM-PoseNet \cite{walch2017pose-lstm}, VidLoc \cite{clark2017vidloc}, and MapNet \cite{brahmbhatt2018mapnet}. Both PoseNet15 \cite{kendall2015posenet} and MapNet \cite{brahmbhatt2018mapnet} are used for comparison on the Oxford RobotCar. Although VidLoc \cite{clark2017vidloc} reported results on the LOOP scene, the training/testing sequences are not provided. 

\subsection{Experiments on the 7-Scenes Dataset}
\label{exp:7scenes}
\textbf{Comparison with Image-based Methods} Table~\ref{tab:7scenes_image} shows the results of methods taking single images as input and our model. Obviously, our approach outperforms these methods by a large margin. As these algorithms rely on single images to recover global poses, their results inevitably suffer from high uncertainties.  

\textbf{Comparison with Sequence-based Methods} Table~\ref{tab:scenes_sequence} demonstrates the comparison against sequence-based relocalization methods including VidLoc~\cite{clark2017vidloc}, MapNet~\cite{brahmbhatt2018mapnet} and DSO~\cite{engel2018dso}. Our approach achieves slightly better performance on regular scenes including \textit{Fire}, \textit{Heads}, \textit{Kitchen}. While on scenarios with many texture-less regions (\textit{Pumpkin}) and highly repetitive textures (\textit{Stairs}), our model obtains outstanding results. As VidLoc \cite{clark2017vidloc} only considers the fusion of observations, thus achieves lower accuracy in translation. MapNet \cite{brahmbhatt2018mapnet} introduces the pose constraints during the training process and achieves promising performance on regular scenes, while our model gives results with higher accuracy. More importantly, our methods shows great potential in dealing with difficult scenarios.

Fig.~\ref{fig:7scenes} illustrates the qualitative comparison between previous methods and our approach.  As PoseNet and its variations~\cite{kendall2015posenet, kendall2016modelling, kendall2017geometric} localize the camera from single images individually, they produce many outliers in both regular and challenging scenes. The number of outliers are reduced by introducing motion constraints over outputs as MapNet \cite{brahmbhatt2018mapnet}. However, this strategy brings finite improvements on scenes with texture-less regions and repetitive textures, resulting in zigzag trajectories. By considering the spatio-temporal consistency of sequential images, our model gains much smoother trajectories in the \textit{Pumpkin} and \textit{Stairs} categories. 

\setlength{\tabcolsep}{3.pt}
\begin{table}[t]
	\footnotesize
	\centering
	\setlength{\abovecaptionskip}{0pt}%
	\setlength{\belowcaptionskip}{0pt}%
		\begin{center}
			\begin{tabular}{lcccc}
				\hline
				\hline
				& \multicolumn{4}{c}{Method} \\
				Scene & DSO~-\cite{engel2018dso} & VidLoc \cite{clark2017vidloc} & MapNet \cite{brahmbhatt2018mapnet}  & \textbf{Ours} \\
				\hline
				
				Chess &0.17m, $8.13^\circ$ & 0.18m, NA &  \textbf{0.08}m, $\mathbf{3.25}^\circ$ & 0.09m, $3.28^\circ$ \\
				Fire & 0.19m, $65.0^\circ$ &0.26m, NA & 0.27m, $11.69^\circ$ & \textbf{0.26}m, $\mathbf{10.92}^\circ$ \\
				Heads & 0.61m, $68.2^\circ$ &\textbf{0.14}m, NA & 0.18m, $13.25^\circ$ & 0.17m, $\mathbf{12.70}^\circ$ \\
				Office & 1.51m, $16.8^\circ$ & 0.26m, NA & \textbf{0.17}m, $\mathbf{5.15}^\circ$ & 0.18m, $5.45^\circ$ \\
				Pumpkin & 0.61m, $15.8^\circ$ & 0.36m, NA & 0.22m, $4.02^\circ$ & \textbf{0.20}m, $\mathbf{3.69}^\circ$ \\
				Kitchen & 0.23m, $10.9^\circ$ & 0.31m, NA & 0.23m, $4.93^\circ$ & \textbf{0.23}m, $\mathbf{4.92}^\circ$ \\
				Stairs & 0.26m, $21.3^\circ$ & 0.26m, NA & 0.30m, $12.08^\circ$ & \textbf{0.23}m, $\mathbf{11.3}^\circ$\\
				Avg & 0.26m, $29.4^\circ$ & 0.25m, NA & 0.21m, $7.77^\circ$ & \textbf{0.19}m, $\mathbf{7.47}^\circ$\\

				\hline
				\hline
			\end{tabular}
		\end{center}
	
	\caption{Median translation and rotation errors of DSO~\cite{engel2018dso}, VidLoc \cite{clark2017vidloc}, MapNet \cite{brahmbhatt2018mapnet} and  our approach on the 7-Scenes dataset \cite{shotton2013scene}. The best results are highlighted.}
	\label{tab:scenes_sequence}
\end{table}
\setlength{\tabcolsep}{1.4pt}

\subsection{Experiments on the Oxford RobotCar Dataset}
\label{exp:robotcar}
\begin{figure}[t]
	\def\subfig_width{0.16}
\centering
\begin{subfigure}{\subfig_width\textwidth}
	\includegraphics[width=0.95\textwidth]{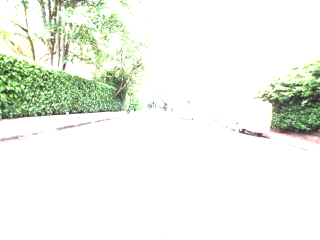}
\end{subfigure}%
\begin{subfigure}{\subfig_width\textwidth}
	\includegraphics[width=0.95\textwidth]{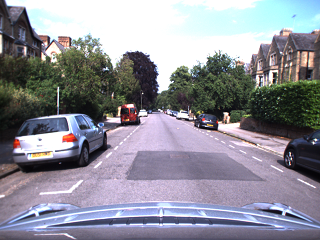}
\end{subfigure}%
\begin{subfigure}{\subfig_width\textwidth}
	\includegraphics[width=0.95\textwidth]{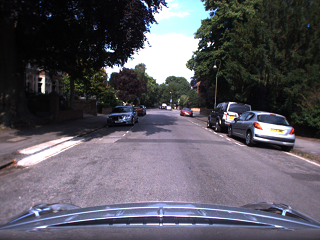}
\end{subfigure}%

\begin{subfigure}{\subfig_width\textwidth}
	\includegraphics[width=0.95\textwidth]{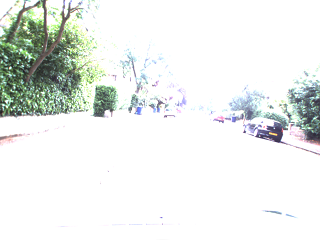}
\end{subfigure}%
\begin{subfigure}{\subfig_width\textwidth}
	\includegraphics[width=0.95\textwidth]{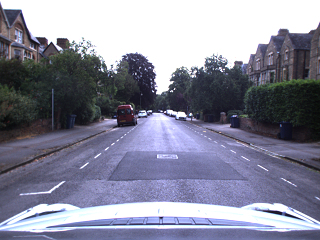}
\end{subfigure}%
\begin{subfigure}{\subfig_width\textwidth}
	\includegraphics[width=0.95\textwidth]{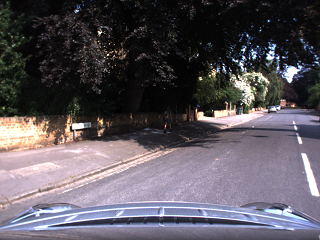}
\end{subfigure}
\caption{Samples of challenging conditions on the RobotCar dataset \cite{robotcar2017}. From left to right: three pairs of images demonstrate the observations with over-exposure, changing weather and dynamic objects, similar appearances at different locations.}
\label{fig:challening_samples}
\end{figure}

We evaluate our model on the LOOP and FULL routes of the Oxford RobotCar dataset \cite{robotcar2017}. Both of the two routes are very long with trajectories of approximate 1120m and 9562m. The training and testing sequences are captured under various weather conditions, which makes them laborious for localization methods. Other uncertainties including dynamic objects, similar appearances, and over exposure further increase the difficulty, as shown in Fig.~\ref{fig:challening_samples}.

\textbf{Quantitative Comparison} 
Table~\ref{tab:error_robotcar} shows the quantitative comparison of our approach against PoseNet and MapNet. Since the training and testing sequences are captured under varying conditions at different times, PoseNet can hardly handle such changes and outputs poor poses. MapNet produces promising results on shorter routes (LOOP1 and LOOP2), but the error increases sharply with the scene growing larger (from 1120m to 9562m). The larger areas may contain more locally similar appearances, degrading the ability the relocalization systems. In contrast, taking both the content and motion into consideration, our model deals with these challenges more effectively.

\setlength{\tabcolsep}{1.4pt}

\begin{table}[t]
	\setlength{\abovecaptionskip}{0pt}%
	\setlength{\belowcaptionskip}{0pt}%
	\centering
		\begin{center}
			\begin{tabular}{lccc}
				\hline
				\hline
				& \multicolumn{3}{c}{Method} \\
				Scene &  PoseNet \cite{kendall2015posenet, kendall2016modelling, kendall2017geometric} & MapNet \cite{brahmbhatt2018mapnet} & \textbf{Ours} \\
				LOOP1 & 28.81m, $19.62^\circ$ & \textbf{8.76}m, $3.46^\circ$ & 9.07m, $\mathbf{3.31}^\circ$  \\
				LOOP2 & 25.29m, $17.45^\circ$ & 9.84m, $3.96^\circ$ &\textbf{9.19}m, $\mathbf{3.53}^\circ$ \\
				FULL1 & 125.6m, $27.1^\circ$ &41.4m, $12.5^\circ$ &\textbf{31.65}m, $\mathbf{4.51}^\circ$ \\
				FULL2 & 131.06m, $26.05^\circ$ &59.30m, $14.81^\circ$ &\textbf{53.45}m, $\mathbf{8.60}^\circ$ \\
				Avg & 77.85m, $22.56^\circ$ & 29.83m, $8.68^\circ$ & \textbf{25.84}m, $\mathbf{4.99}^\circ$ \\
				
				
				\hline
				\hline
			\end{tabular}
		\end{center}
	
	\caption{Mean translation and rotation errors of PoseNet \cite{kendall2015posenet, kendall2016modelling, kendall2017geometric}, MapNet \cite{brahmbhatt2018mapnet} nad our method on Oxford RobotCar dataset \cite{robotcar2017}. Results of PoseNet and MapNet are from \cite{brahmbhatt2018mapnet}. The best results are highlighted.}
	\label{tab:error_robotcar}
\end{table}

\begin{figure*}[t]
	
\def\subfig_width{0.25}
\centering
\begin{subfigure}{\subfig_width\textwidth}
	\includegraphics[width=1\textwidth]{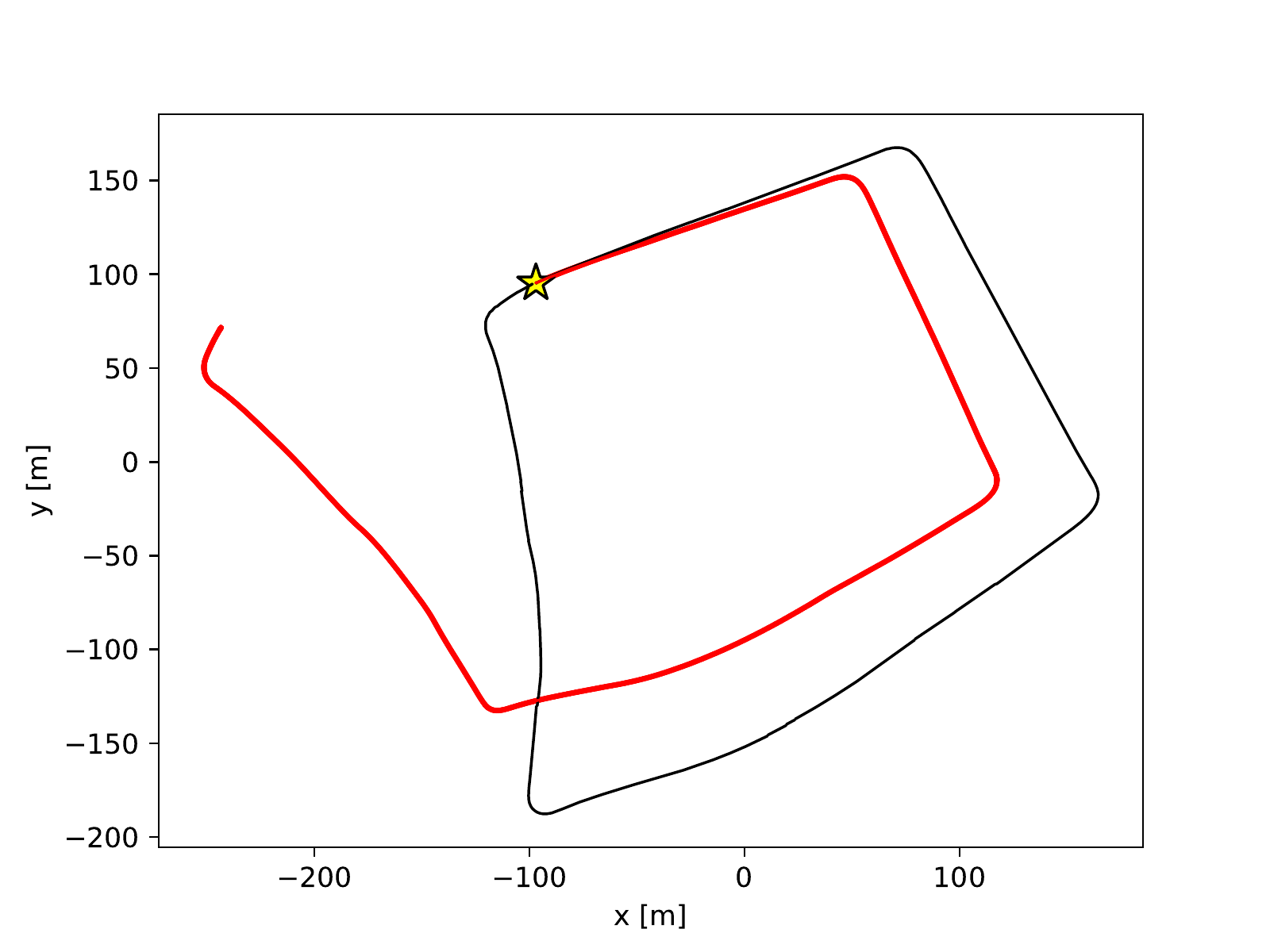}
\end{subfigure}%
\begin{subfigure}{\subfig_width\textwidth}
	\includegraphics[width=1\textwidth]{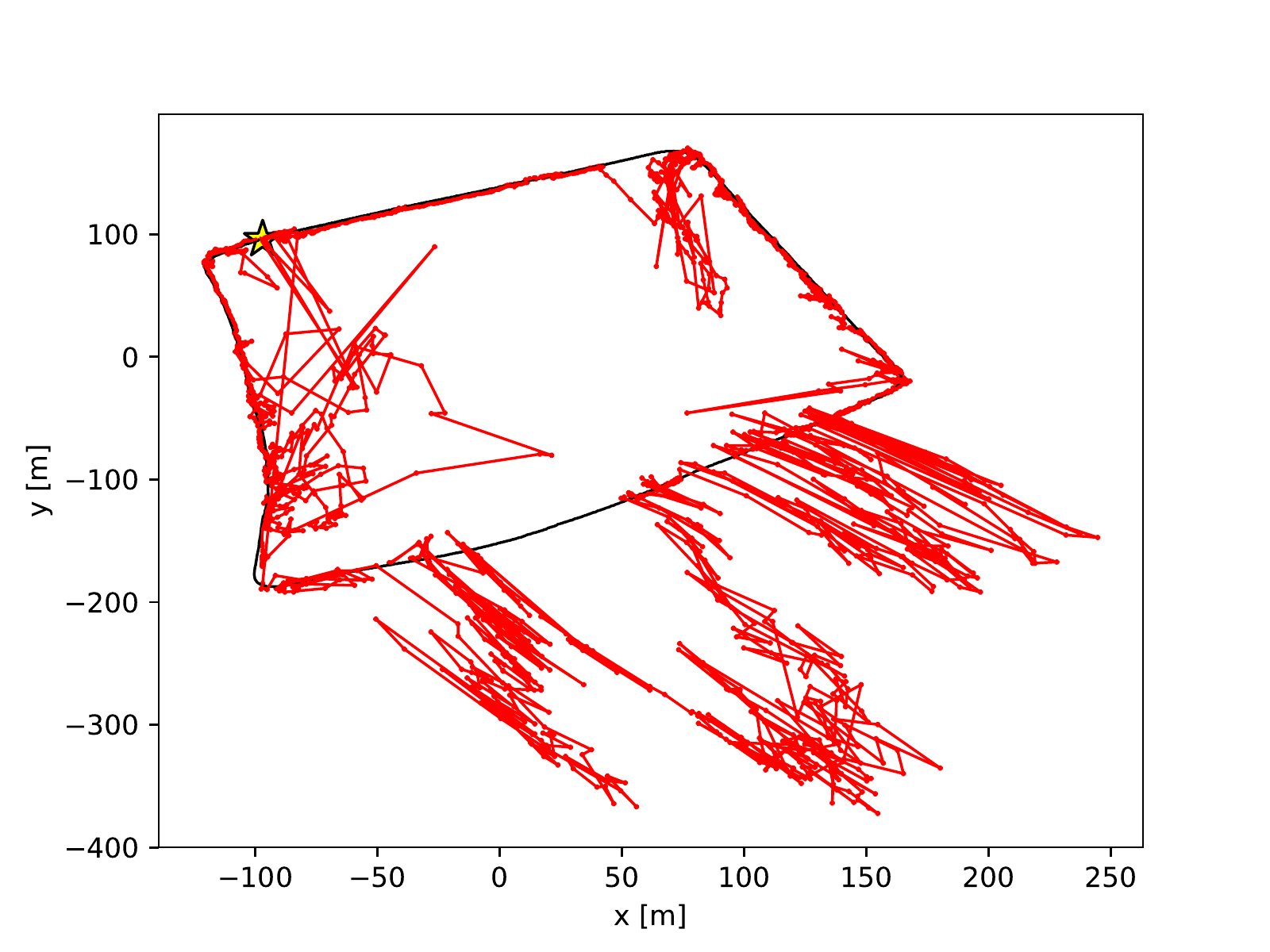}
\end{subfigure}%
\begin{subfigure}{\subfig_width\textwidth}
	\includegraphics[width=1\textwidth]{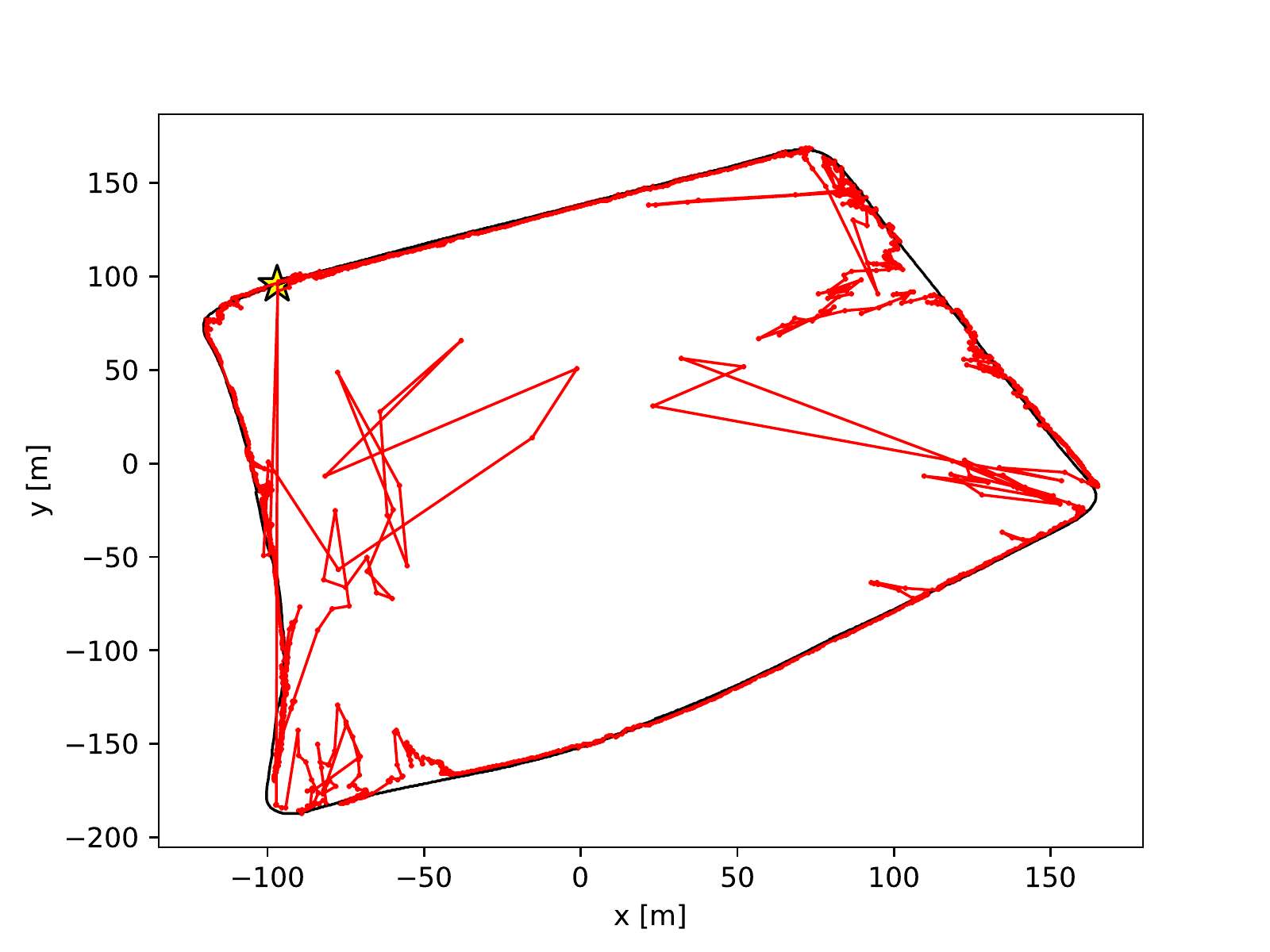}
\end{subfigure}%
\begin{subfigure}{\subfig_width\textwidth}
	\includegraphics[width=1\textwidth]{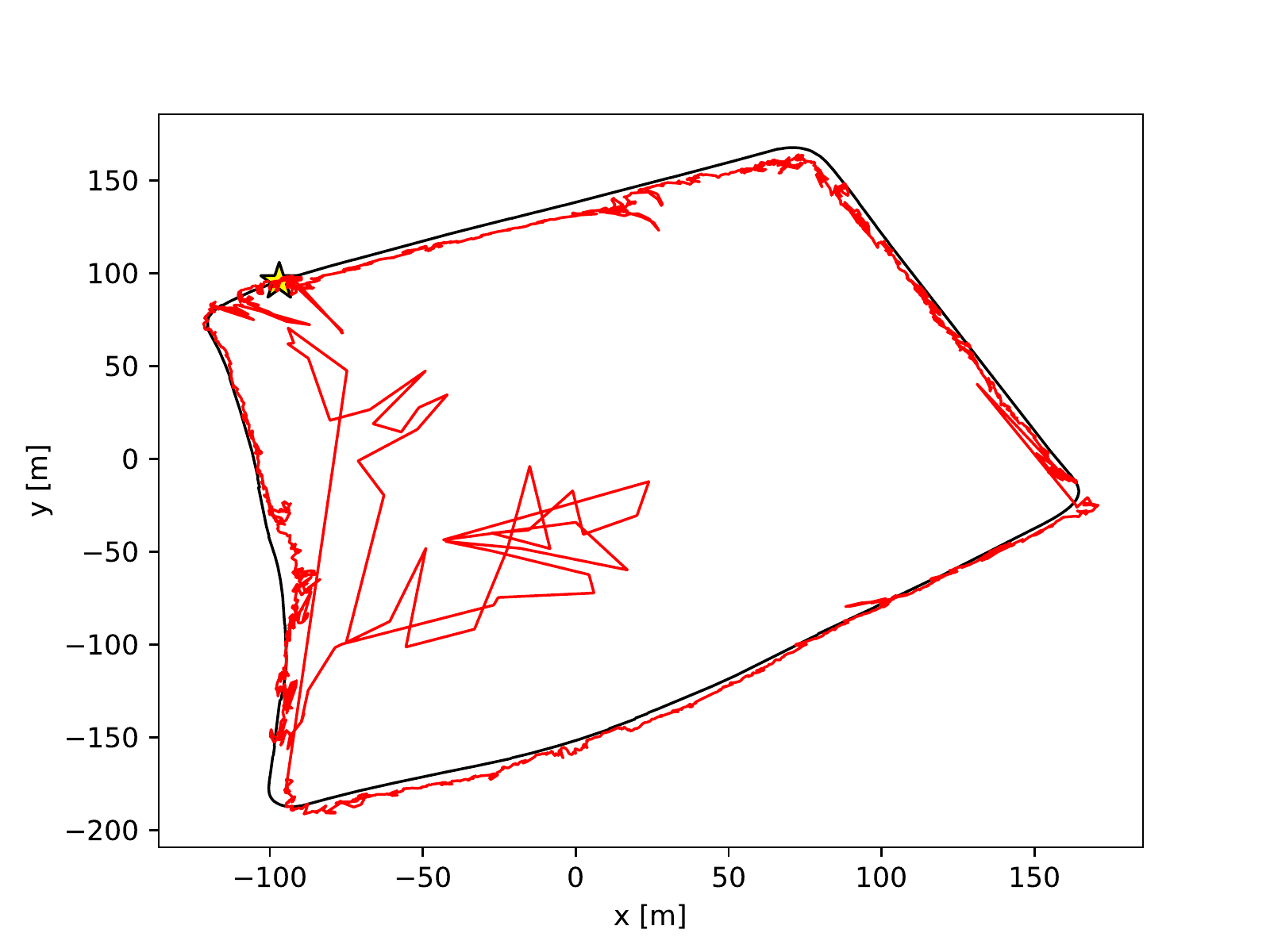}
\end{subfigure}%

\begin{subfigure}{\subfig_width\textwidth}
	\includegraphics[width=1\textwidth]{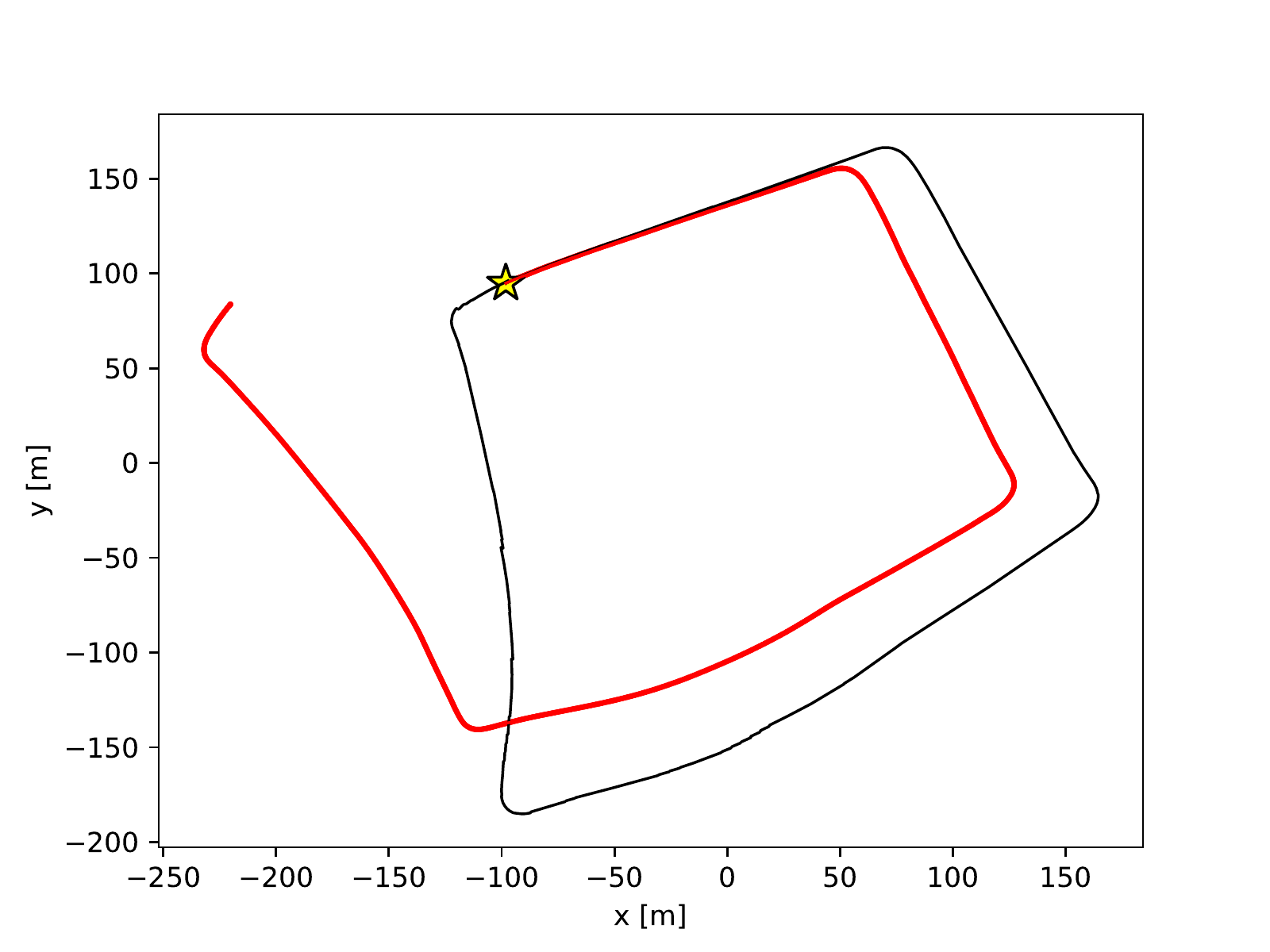}
\end{subfigure}%
\begin{subfigure}{\subfig_width\textwidth}
	\includegraphics[width=1\textwidth]{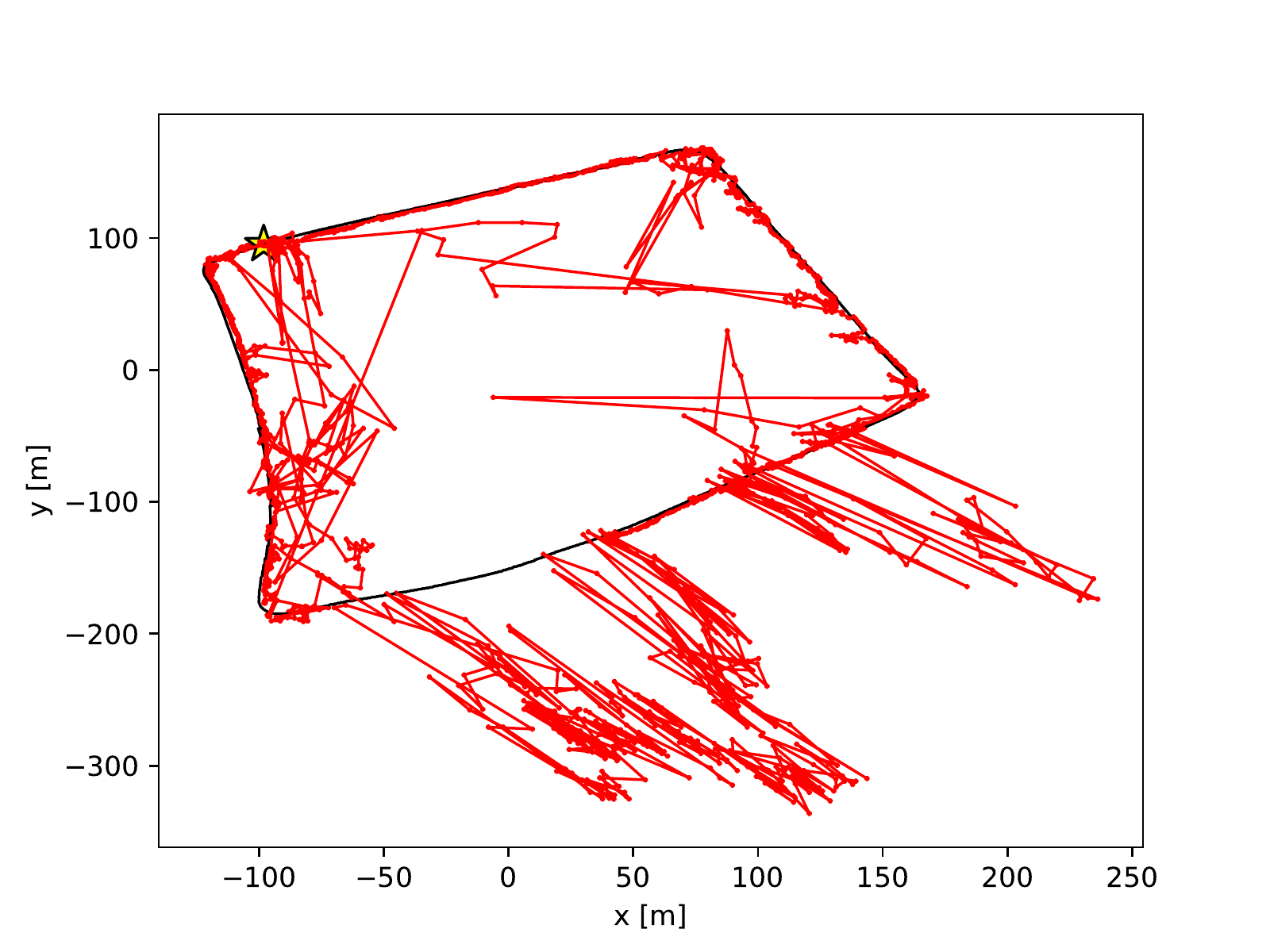}
\end{subfigure}%
\begin{subfigure}{\subfig_width\textwidth}
	\includegraphics[width=1\textwidth]{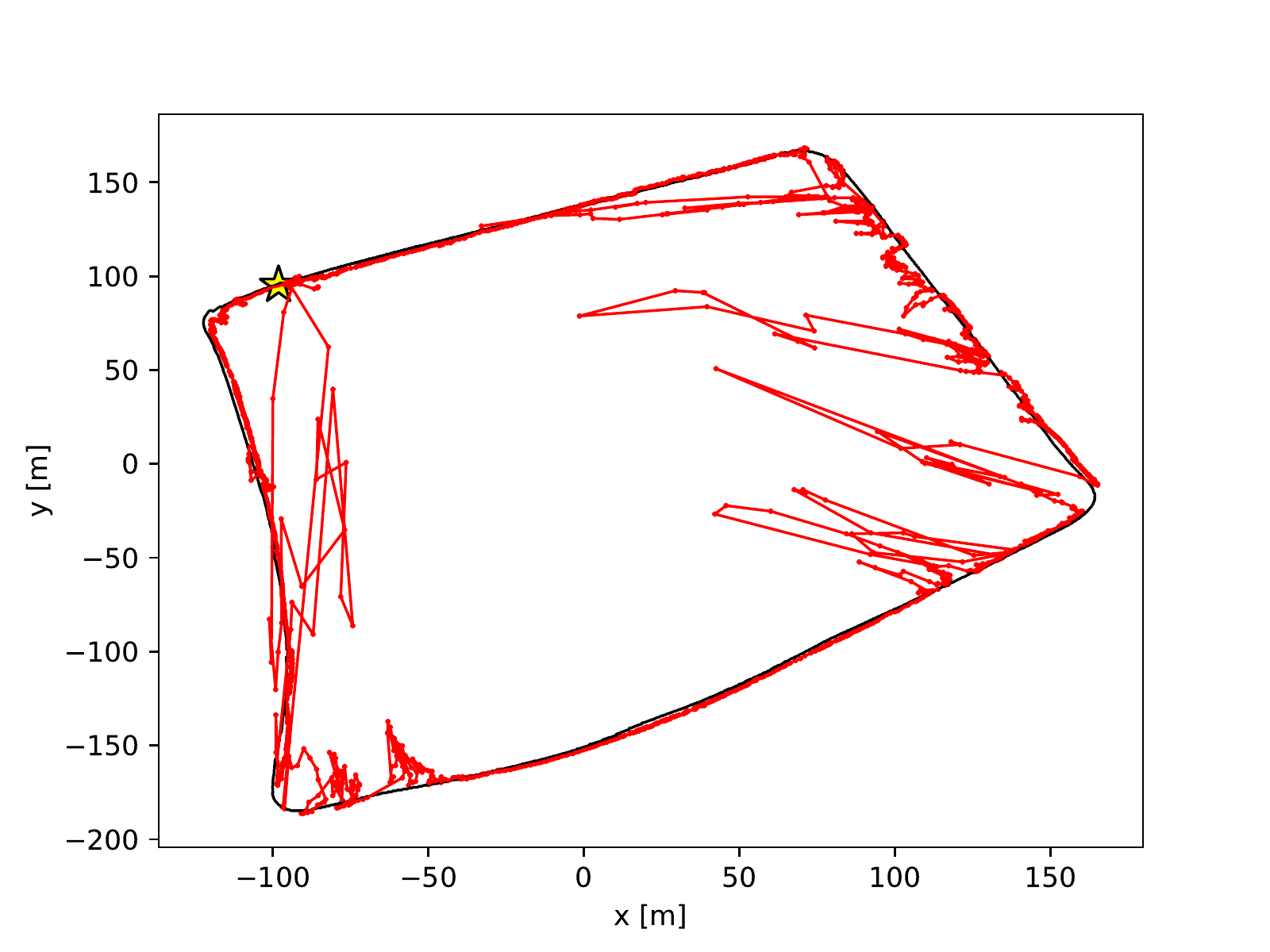}
\end{subfigure}%
\begin{subfigure}{\subfig_width\textwidth}
	\includegraphics[width=1\textwidth]{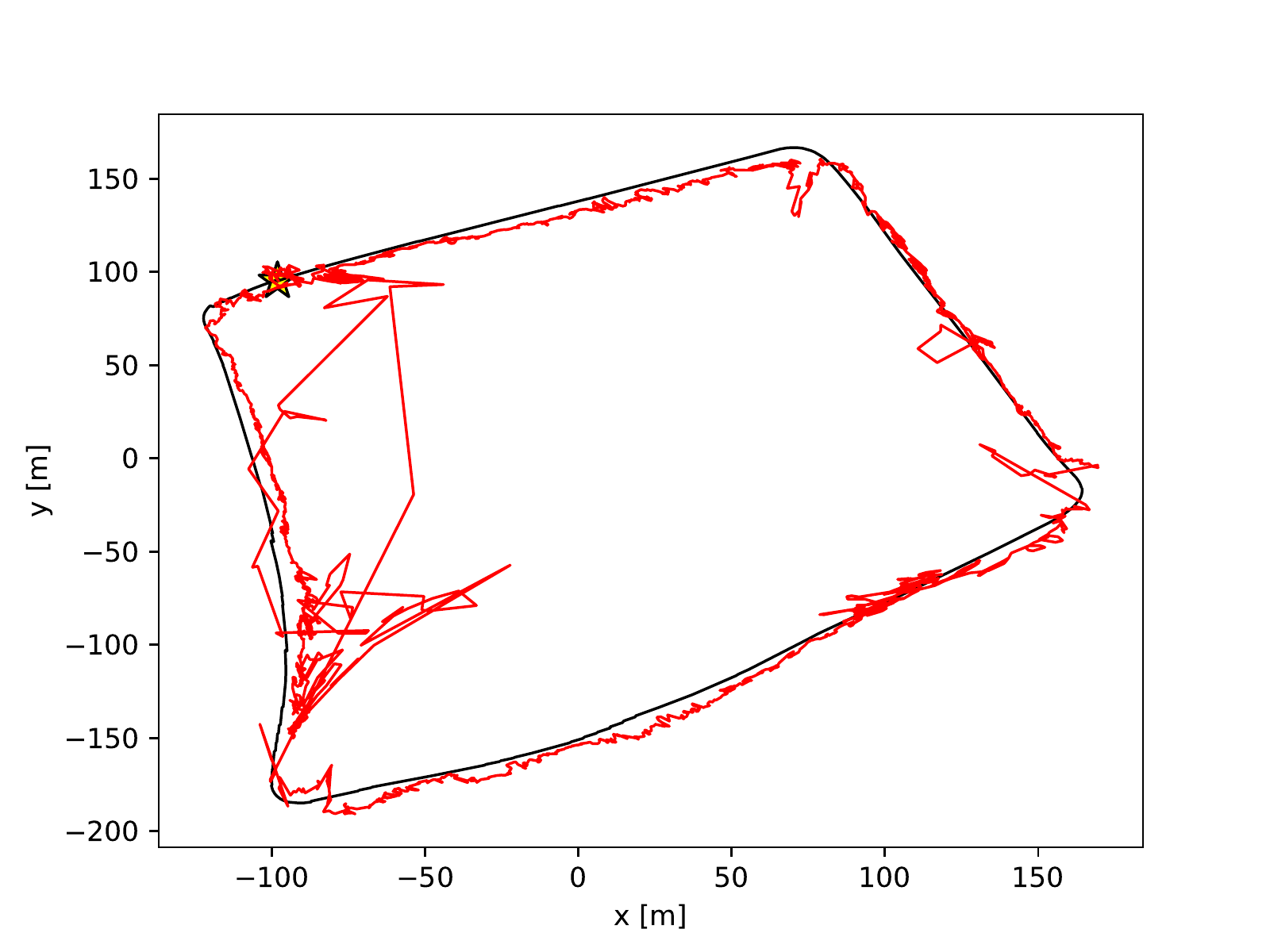}
\end{subfigure}%

\begin{subfigure}{\subfig_width\textwidth}
	\includegraphics[width=1\textwidth]{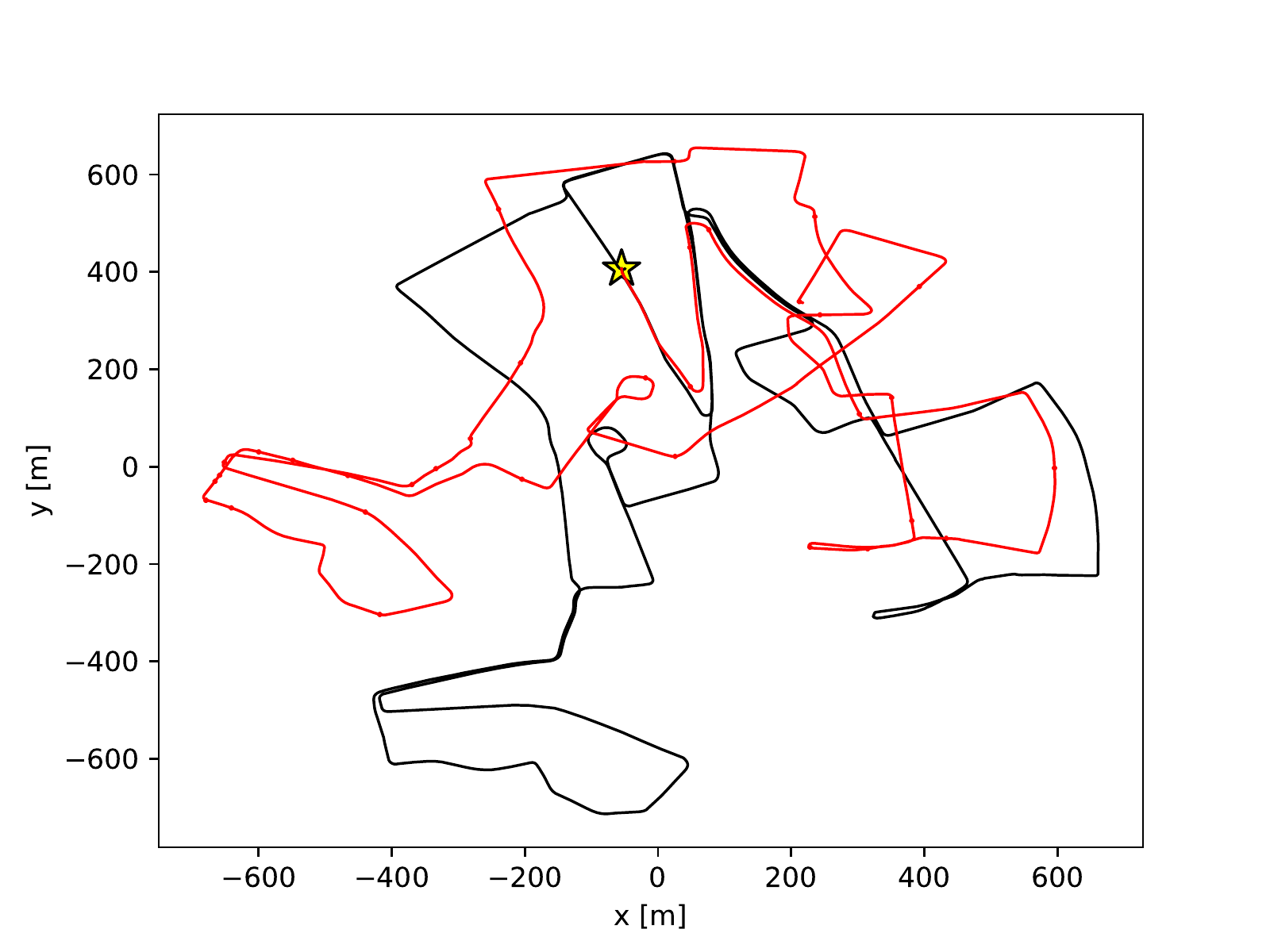}
	\centerline{Stereo VO \cite{robotcar2017}}
\end{subfigure}%
\begin{subfigure}{\subfig_width\textwidth}
	\includegraphics[width=1\textwidth]{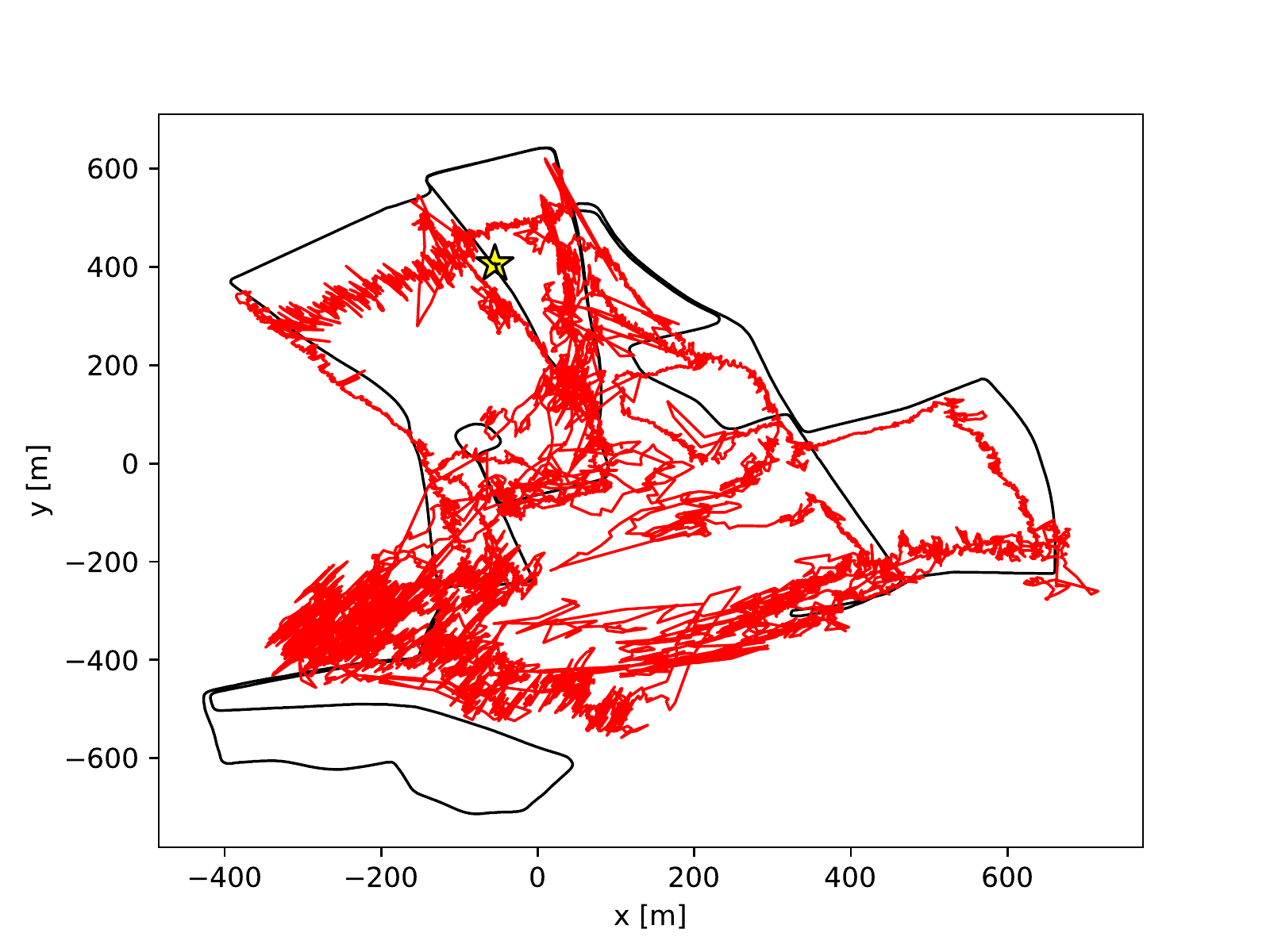}
	\centerline{PoseNet \cite{kendall2015posenet, kendall2016modelling, kendall2017geometric}}
\end{subfigure}%
\begin{subfigure}{\subfig_width\textwidth}
	\includegraphics[width=1\textwidth]{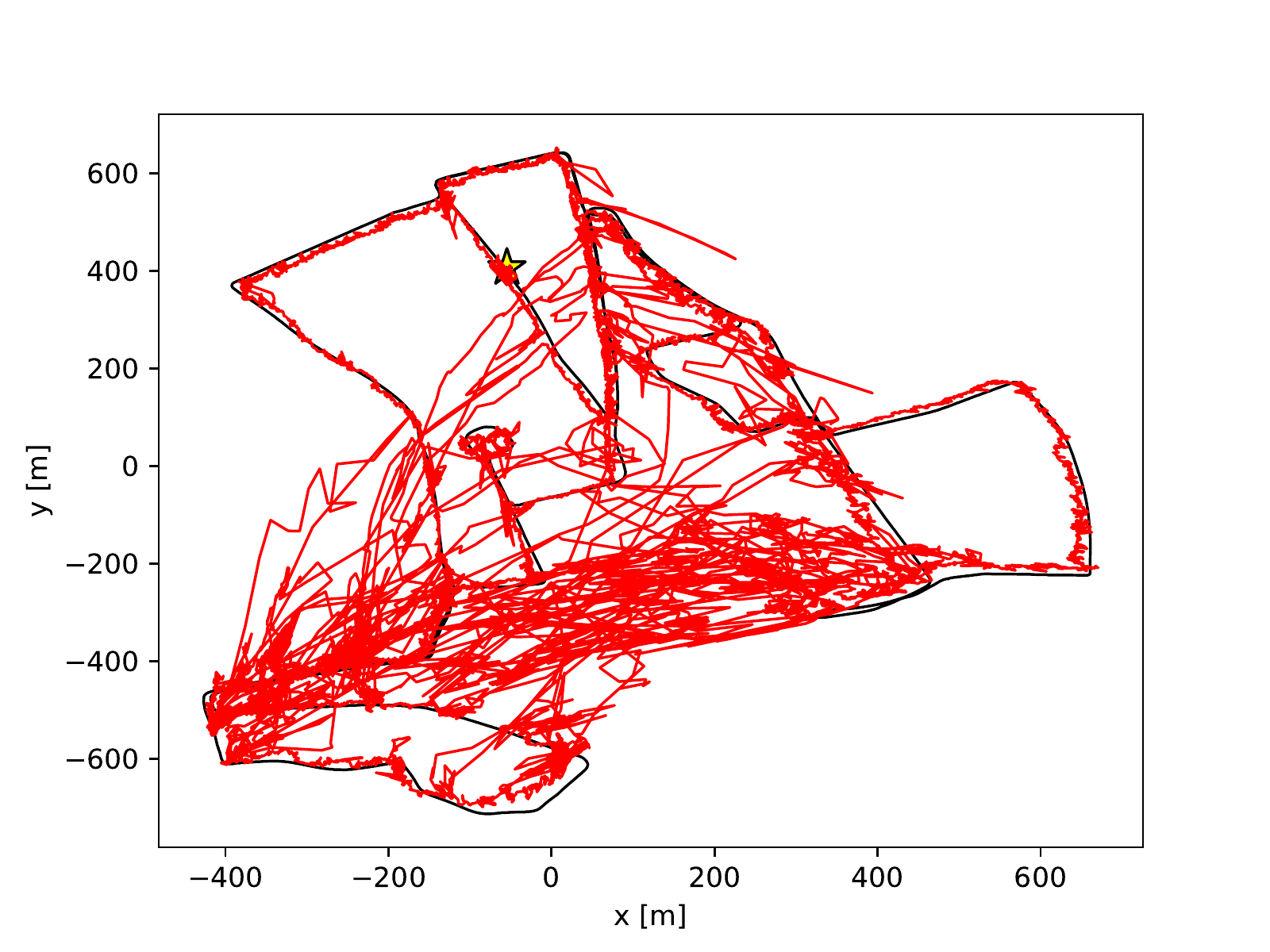}
	\centerline{MapNet \cite{brahmbhatt2018mapnet}}
\end{subfigure}%
\begin{subfigure}{\subfig_width\textwidth}
	\includegraphics[width=1\textwidth]{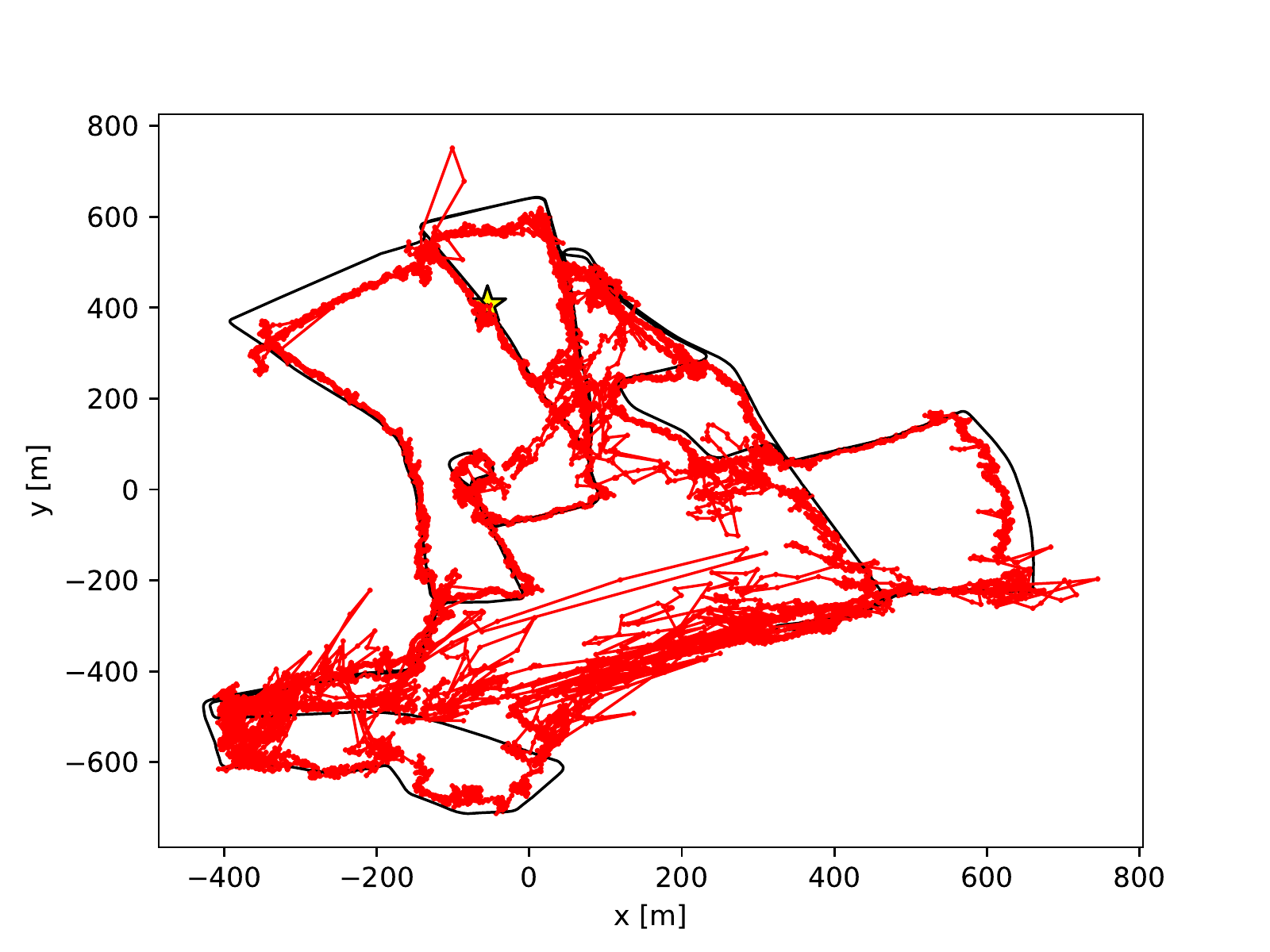}
	\centerline{Ours}
\end{subfigure}
	\caption{Results of Stereo VO, PoseNet, MapNet and our model on the LOOP1 (top), LOOP2 (middle) and FULL1 scenes (bottom) of the Oxford RobotCar dataset \cite{robotcar2017} The red and black lines indicate predicted and ground truth poses respectively. }
	\label{fig:exp_robotcar_loop}
\end{figure*}
\setlength{\tabcolsep}{1.4pt}

\begin{figure*}[t]
	\def\subfig_width{0.25}
\centering
\begin{subfigure}{\subfig_width\textwidth}
	\includegraphics[width=1\textwidth]{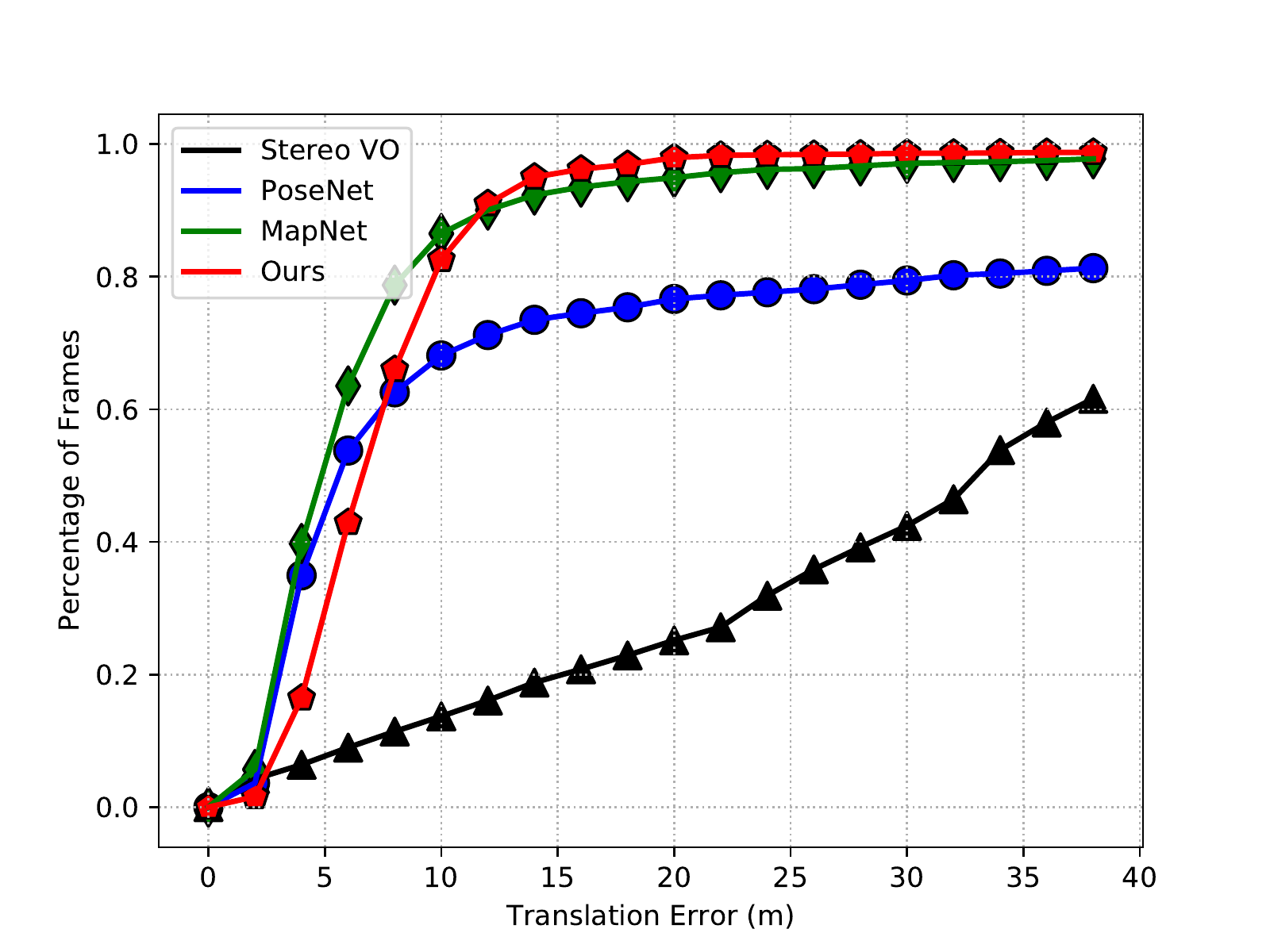}
	\caption{Translation error on LOOP.}
\end{subfigure}%
\begin{subfigure}{\subfig_width\textwidth}
	\includegraphics[width=1\textwidth]{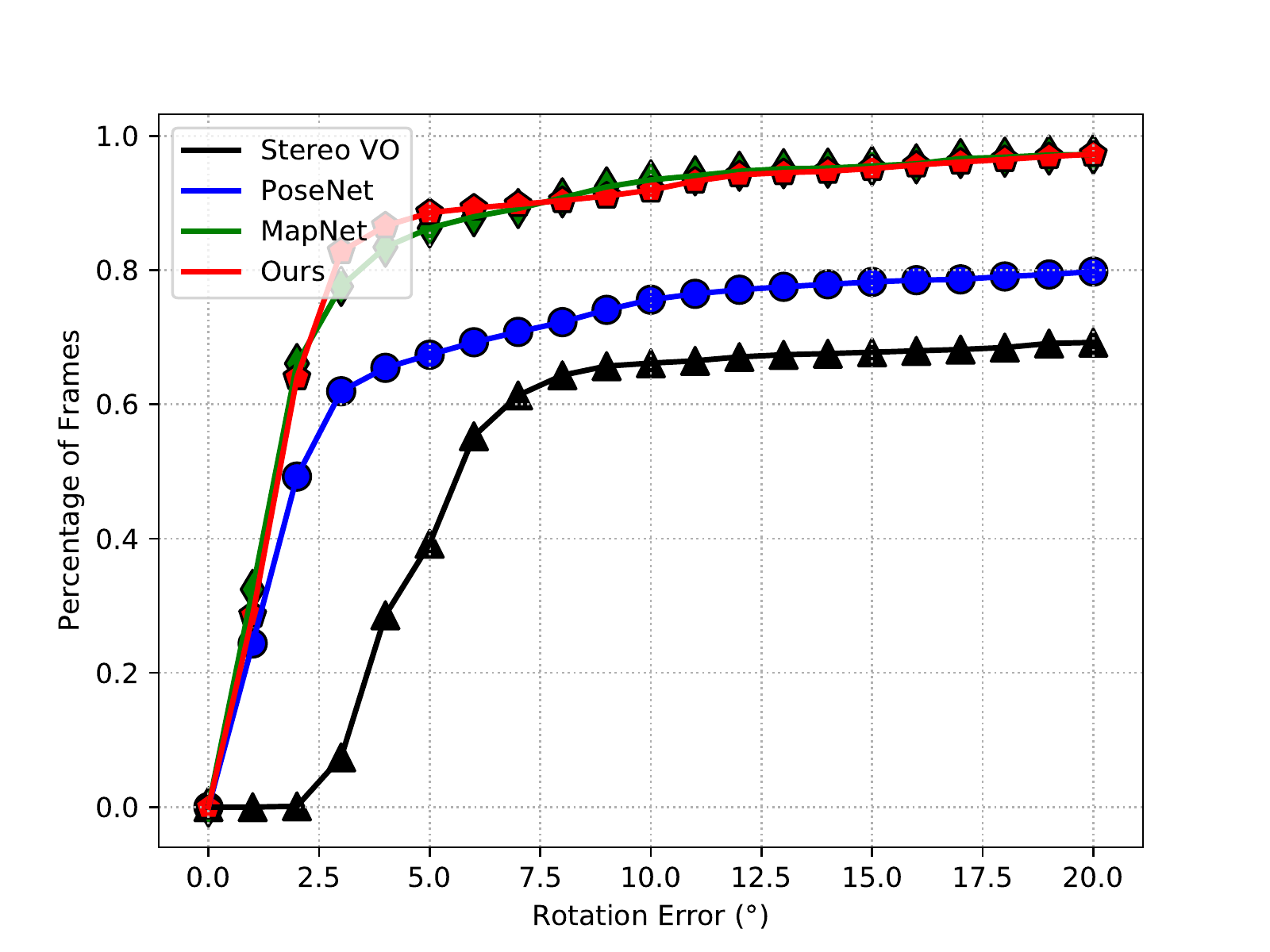}
	\caption{Rotation error on LOOP.}
\end{subfigure}%
\begin{subfigure}{\subfig_width\textwidth}
	\includegraphics[width=1\textwidth]{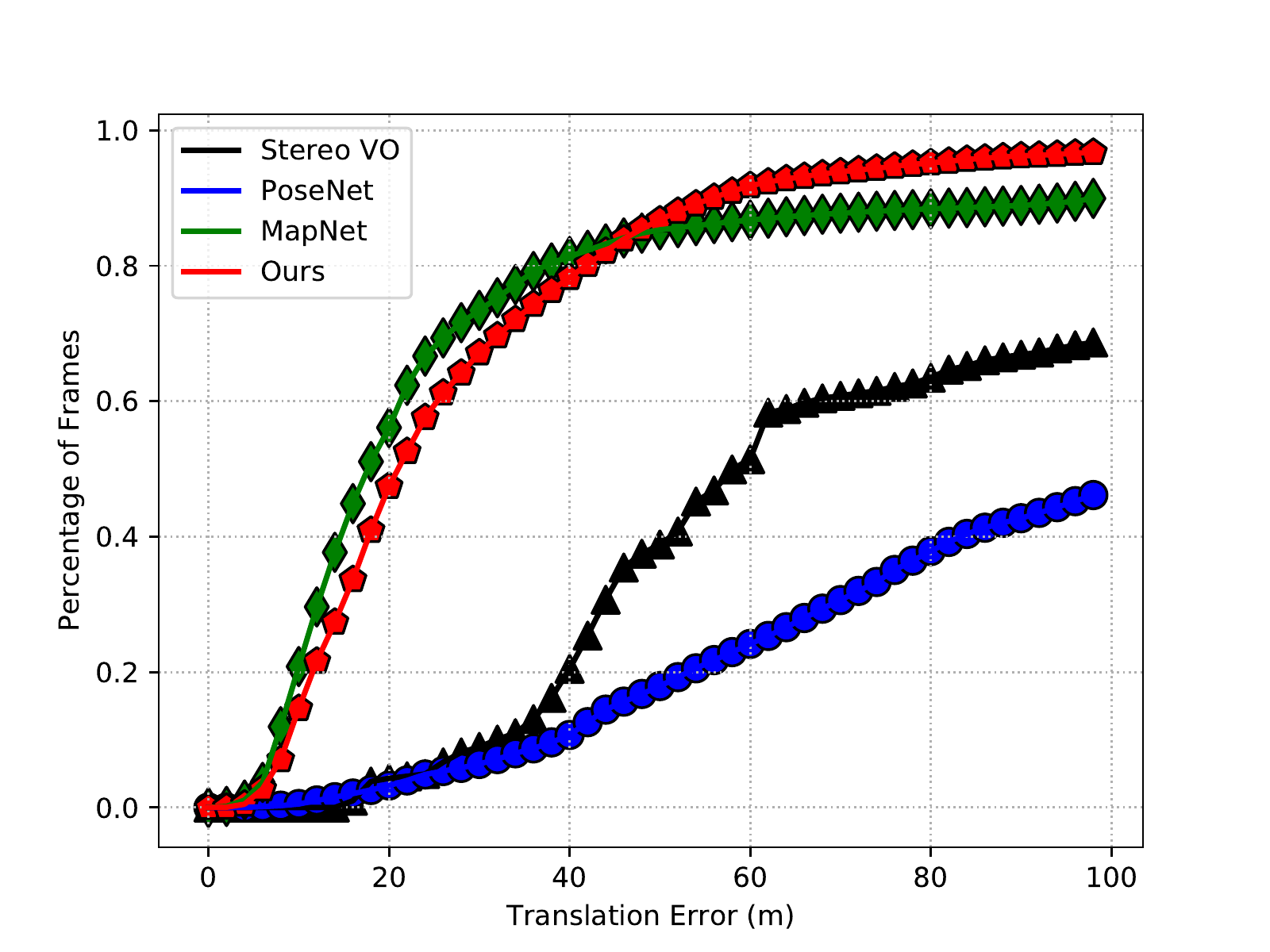}
	\caption{Translation error on FULL.}
\end{subfigure}%
\begin{subfigure}{\subfig_width\textwidth}
	\includegraphics[width=1\textwidth]{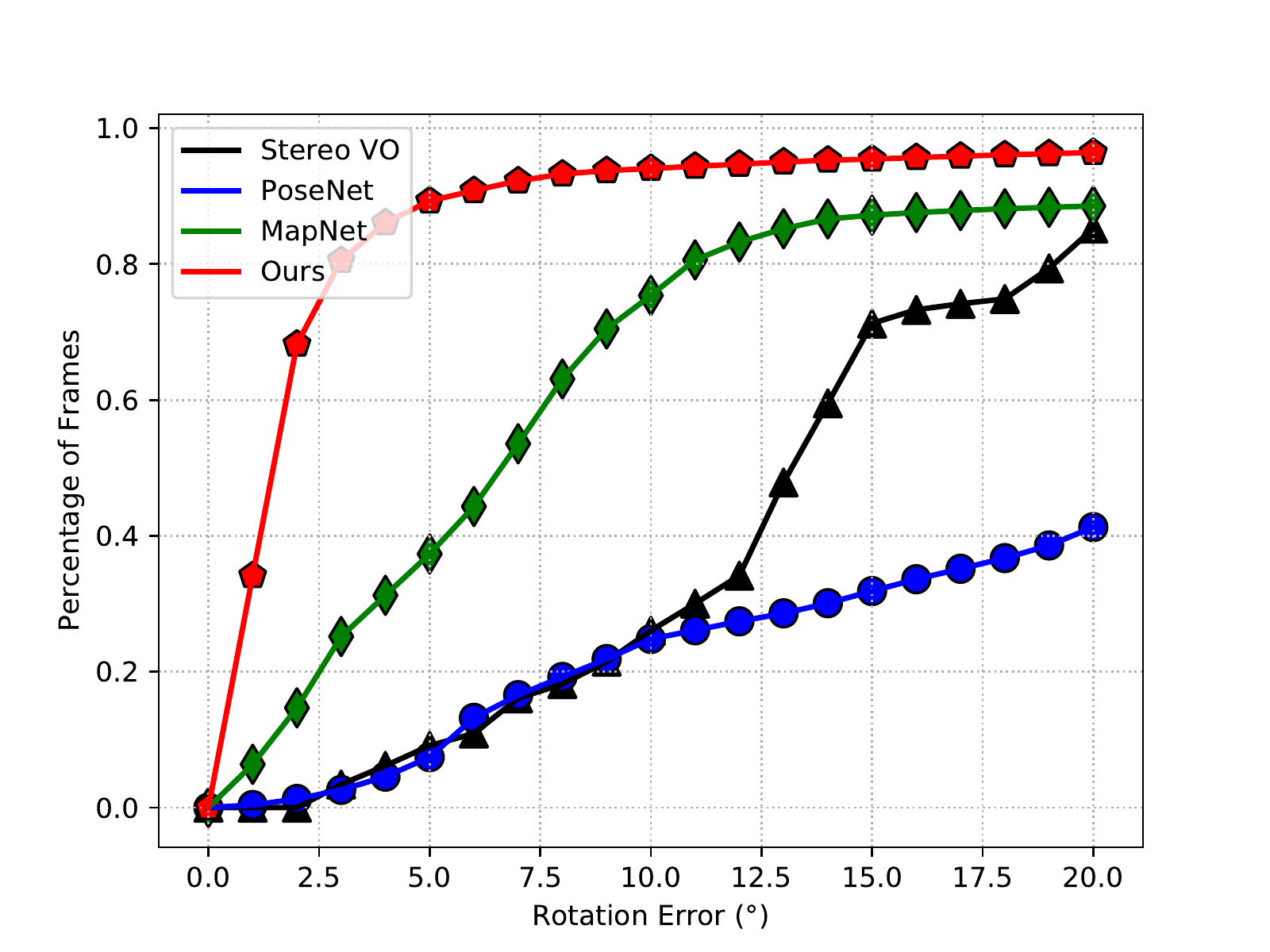}%
	\caption{Rotation error on FULL.}
\end{subfigure}
	\caption{Cumulative distributions of the mean translation (m) and rotation errors ($\circ$) of all the methods evaluated on Oxford RobotCar LOOP and FULL scenes. X-axis is the errors and y-axis is the percentage of frames with error less than the value. }
	\label{fig:exp_robotcar_error}
\end{figure*}
\setlength{\tabcolsep}{1.4pt}

\textbf{Qualitative Comparison} Fig.~\ref{fig:exp_robotcar_loop} represents the trajectories recovered by Stereo VO (results from \cite{robotcar2017}), PoseNet, MapNet and our model. As shown clearly, Stereo VO produces very smooth trajectories as the relative poses between consecutive frames can be calculated locally accurately. Yet, Stereo VO drifts severely over time in such long routes. PoseNet predicts a lot of outliers which are far from the ground truth locations. Utilizing the motion consistency, MapNet \cite{brahmbhatt2018mapnet} improves the accuracy of predicted poses and reduces the number of outliers by a large margin. However, there are still a lot of outliers across the whole areas due to the local similarities of observations in large scenarios. Compared with PoseNet and MapNet, our model eliminates outliers more effectively and performs more stably in these areas where the previous two methods give noisy results.

\textbf{Cumulative Distribution Errors} We additionally calculate the cumulative distribution of the translation and rotation errors on the LOOP and FULL scenes (shown in Fig.~\ref{fig:exp_robotcar_error}). We can find that Stereo VO and PoseNet produce poor results in both translation and rotation. MapNet and our method produce very close results in regular regions. While for scenarios causing larger uncertainties, the performance of our model is markedly more stable. Deep learning techniques are not good at estimating camera orientations \cite{Sattler_2018_CVPR}, let alone from single images. Therefore, another important advantage of our model is its ability to give accurate orientation estimation by considering the local information of sequential images.





\section{Conclusion}
\label{conclusion}

In this paper, we aim to overcome the ambiguities of single images in relocalization with assistance of the local information in image sequences. To fully use the spatio-temporal consistency, we incorporate a VO component. Specifically, instead of raw features, our model estimates global poses from features which are augmented by local maps preserved in recurrent units based on the co-visibility. Besides, as relative poses can be calculated with promising accuracy in VO component due to the continuity of camera motions, the predicted global poses are further optimized with these relative poses constraints within a pose graph during both the training and testing processes. Experiments on both the indoor 7-Scenes and outdoor Oxford RobotCar datasets demonstrate that our approach outperforms previous methods, especially in challenging scenarios. 

\section*{Acknowledgement}
The work is supported by the National Key Research and Development Program of China (2017YFB1002601) and National Natural Science Foundation of China (61632003, 61771026).

{\small
\bibliographystyle{ieee}
\bibliography{egbib}
}

\newpage





\begin{abstract}
\label{sec:abs}
In the supplementary material, we first introduce the training and testing sequences adopted on the Oxford RobotCar dataset \cite{robotcar2017} in Table~\ref{tab:train_test_split}. The descriptions of corresponding sequences are also included. 

Moreover, we perform an ablation study in Sec.~\ref{sec:ablation_study}. In Sec.~\ref{sec:robotcar}, additional comparisons against previous methods on various challenging scenes of the Oxford RobotCar dataset are presented. We visualize the attention maps produced by PoseNet \cite{kendall2015posenet, kendall2016modelling, kendall2017geometric}, MapNet \cite{brahmbhatt2018mapnet} and our model in Sec.~\ref{sec:visualization}
\end{abstract}

\section*{Ablation Study}
\label{sec:ablation_study}
We perform an ablation study to evaluate the effectiveness of each part of our architecture in Table~\ref{tab:error_robotcar_ablation}. Our basic model achieves the poorest performance because both the content augmentation and motion constraints are disabled. Both the translation and orientation errors are reduced by introducing the content augmentation. The performance is further enhanced by adding the motion constraints.


\section*{Robustness to Various Conditions}
\label{sec:robotcar}

\begin{figure*}[t]
	\def\subfig_width{0.2}
\centering
\begin{subfigure}{\subfig_width\textwidth}
	\includegraphics[width=0.95\textwidth]{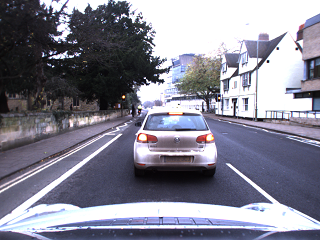}
	\subcaption{December, overcast}
\end{subfigure}%
\begin{subfigure}{\subfig_width\textwidth}
	\includegraphics[width=0.95\textwidth]{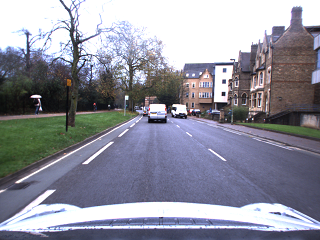}
	\subcaption{November, rain}
\end{subfigure}%
\begin{subfigure}{\subfig_width\textwidth}
	\includegraphics[width=0.95\textwidth]{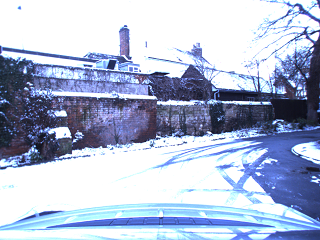}
	\subcaption{February, snow}
\end{subfigure}%
\begin{subfigure}{\subfig_width\textwidth}
	\includegraphics[width=0.95\textwidth]{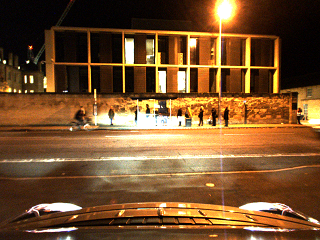}
	\subcaption{December, night}
\end{subfigure}%
\begin{subfigure}{\subfig_width\textwidth}
	\includegraphics[width=0.95\textwidth]{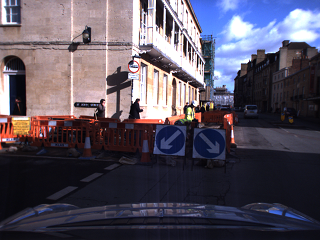}
	\subcaption{February, roadworks}
\end{subfigure}%

\caption{Samples of images captured under different weather, season, illumination, and roadwork conditions on the Oxford RobotCar dataset \cite{robotcar2017}. }
\label{fig:challening_samples_extend}
\end{figure*}

Since practical visual localization approaches need to deal with various conditions, we additionally compare the robustness of PoseNet \cite{kendall2015posenet, kendall2016modelling, kendall2017geometric}, MapNet \cite{brahmbhatt2018mapnet} and our method in handling situations including weather and seasonal variations, as well as day-night changes. Samples of images can be seen in Fig.~\ref{fig:challening_samples_extend}. As Oxford RobotCar dataset \cite{robotcar2017} provides sequences fulfilling these conditions, we test the generalization ability of three models quantitatively and qualitatively. It's worthy to note that our model is trained on the same sequences as PoseNet and MapNet, without any fine-tuning.

\subsection*{Quantitative Comparison}
Table~\ref{tab:error_robotcar_extend} presents the adopted sequences, descriptions and quantitative results. Performances of both PoseNet and MapNet degrade a lot due to the challenging changes. We notice that MapNet achieves very close results with PoseNet. The possible reason is that MapNet takes only singe images to regress global camera poses, as PoseNet. Though MapNet employs motion constraints over several frames during training,  motion constraints are incapable of mitigating visual ambiguities existing in challenging scenarios. In contrast, our content augmentation strategy copes with these problems effectively (see attention maps in Fig.~\ref{fig:vis_samples}) and the pose uncertainties are further alleviated by the motion constraints.

\setlength{\tabcolsep}{3.5pt}
\begin{table}[t]
	\centering
		\begin{center}
			\begin{tabular}{ccccc}
				\hline
				\hline
				Label & Sequence & Tag &  Train & Test \\
				\hline
				-- & 2014-06-26-08-53-56 & overcast&  $\checkmark$ & \\
				-- & 2014-06-26-09-24-58 &  overcast &  $\checkmark$ &  \\
				LOOP1 & 2014-06-23-15-41-25 & sun &  & $\checkmark$\\
				LOOP2 & 2014-06-23-15-36-04 & sun &   & $\checkmark$\\
				\hline
				-- & 2014-11-28-12-07-13 & overcast& $\checkmark$ &\\
				-- & 2014-12-02-15-30-08 & overcast&  $\checkmark$&\\
				FULL1 & 2014-12-09-13-21-02 & overcast&   &$\checkmark$\\
				FULL2 & 2014-12-12-10-45-15 & overcast&  &$\checkmark$\\
				\hline
				\hline
			\end{tabular}
		\end{center}
	
	\caption{Training, testing sequences and corresponding descriptions on the Oxford RobotCar dataset \cite{robotcar2017}. We adopt the same train/test split as PoseNet \cite{kendall2015posenet, kendall2016modelling, kendall2017geometric} and MapNet \cite{brahmbhatt2018mapnet}.}
	\label{tab:train_test_split}
\end{table}
\setlength{\tabcolsep}{1.4pt}
\setlength{\tabcolsep}{3.pt}
\begin{table}[t]
	\small
	\setlength{\abovecaptionskip}{0pt}%
	\setlength{\belowcaptionskip}{0pt}%
	\centering
		\begin{center}
			\begin{tabular}{lccc}
				\hline
				\hline
				& \multicolumn{3}{c}{Method} \\
				Scene & Ours (basic) & Ours (w/ content) & Ours (full) \\
				LOOP1 & 19.39m, $7.56^\circ$ & 9.48m, $4.23^\circ$ & \textbf{9.07}m, $\mathbf{3.31}^\circ$  \\
				LOOP2 & 21.07m, $9.42^\circ$ & 10.56m, $4.37^\circ$ &\textbf{9.19}m, $\mathbf{3.53}^\circ$ \\
				FULL1 & 108.13m, $19.49^\circ$ &59.83m, $10.97^\circ$ &\textbf{31.65}m, $\mathbf{4.51}^\circ$ \\
				FULL2 & 109.73m, $19.01^\circ$ &85.98m, $11.93^\circ$ &\textbf{53.45}m, $\mathbf{8.60}^\circ$ \\
				Avg & 64.58m, $13.87^\circ$ & 41.46m, $7.88^\circ$ & \textbf{25.84}m, $\mathbf{4.99}^\circ$ \\
				
				
				\hline
				\hline
			\end{tabular}
		\end{center}
	
	\caption{Mean translation and rotation errors of variations of our model on the Oxford RobotCar dataset \cite{robotcar2017}. \textbf{Ours (basic)} indicates the model without content augmentation and motion constraints. \textbf{Ours (w/ content)} indicates the model with content augmentation but without motion constraints. \textbf{Ours (full)} contains both the content augmentation and motion constraints. The best results are highlighted.}
	\label{tab:error_robotcar_ablation}
\end{table}
\setlength{\tabcolsep}{1.4pt}
\setlength{\tabcolsep}{1.4pt}

\setlength{\tabcolsep}{5.pt}
\begin{table*}[t]
	\setlength{\abovecaptionskip}{0pt}%
	\setlength{\belowcaptionskip}{0pt}%
	\centering
		\begin{center}
			\begin{tabular}{lcc|ccc}
				\hline
				\hline
				\multicolumn{3}{c|}{Description} & \multicolumn{3}{c}{Method} \\
				\hline
				Scene & Sequence & Tag & PoseNet \cite{kendall2015posenet, kendall2016modelling, kendall2017geometric} & MapNet \cite{brahmbhatt2018mapnet} & \textbf{Ours} \\
				
				FULL3 & 2014-12-05-11-09-10 & overcast, rain & 104.41m, $20.94^\circ$ & 73.74m, $21.06^\circ$ &\textbf{57.54}m, $\mathbf{8.49}^\circ$ \\
				
				FULL4 & 2014-11-25-09-18-32 & overcast, rain &151.24m, $34.70^\circ$ &166.70m, $35.62^\circ$ &\textbf{137.53}m, $\mathbf{23.23}^\circ$ \\
				
				FULL5 & 2015-02-03-08-45-10 & snow & 125.22m, $21.61^\circ$ &139.75m, $29.02^\circ$ &\textbf{71.42}m, $\mathbf{12.92}^\circ$ \\
				
				FULL6 & 2015-02-24-12-32-19 & roadworks, sun &132.86m, $32.22^\circ$ &157.64m, $33.88^\circ$ &\textbf{81.92}m, $\mathbf{16.79}^\circ$ \\
				
				FULL7 & 2014-12-10-18-10-50 & night &405.17m, $75.64^\circ$ &397.80m, $81.40^\circ$ &\textbf{385.58}m, $\mathbf{68.81}^\circ$ \\
				
				FULL8 & 2014-12-17-18-18-43 & night, rain &471.89m, $82.11^\circ$ &430.49m, $85.15^\circ$ &\textbf{430.54}m, $\mathbf{72.35}^\circ$ \\
				Avg & -- & -- & 231.80m, $44.54^\circ$ & 227.69m, $47.69^\circ$ & \textbf{193.98}m, $\mathbf{33.77}^\circ$ \\
				
				
				\hline
				\hline
			\end{tabular}
		\end{center}
	\caption{Mean translation and rotation errors of PoseNet \cite{kendall2015posenet, kendall2016modelling, kendall2017geometric}, MapNet \cite{brahmbhatt2018mapnet} and our method on the Oxford RobotCar dataset \cite{robotcar2017}. Results of PoseNet and MapNet are generated from weights released by \cite{brahmbhatt2018mapnet}. The sequences were captured at different times with day-night changes, as well as weather and seasonal variations. Moreover, changes of traffic, pedestrians, construction and roadworks are also included. The best results are highlighted.}
	\label{tab:error_robotcar_extend}
\end{table*}

\subsection*{Qualitative Comparison}
\begin{figure*}[t]
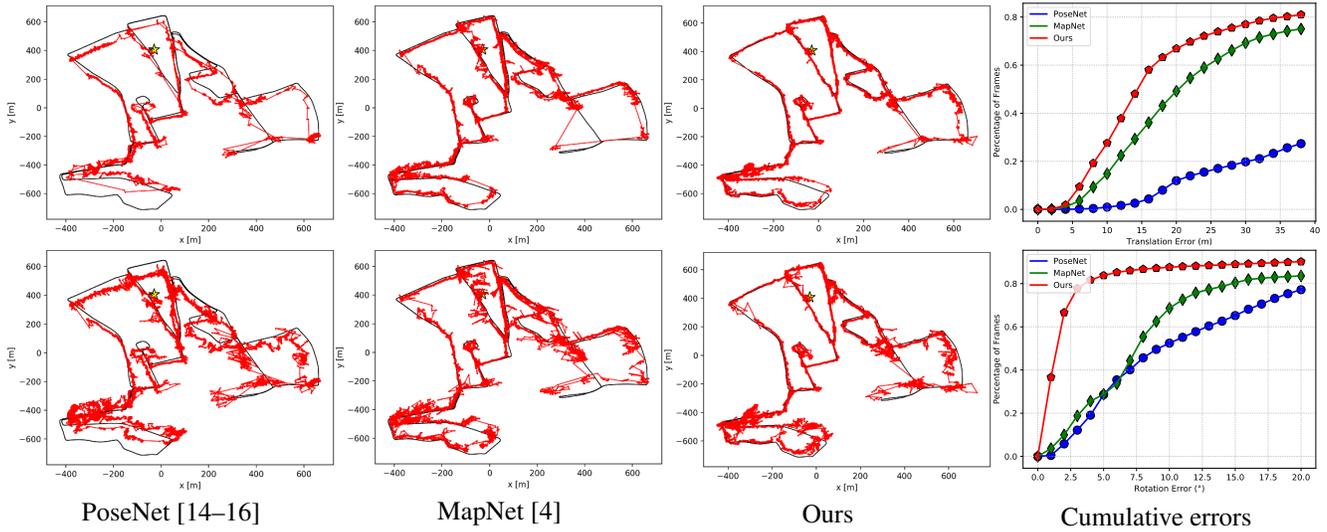

	
\def\subfig_width{0.25}
\def\seq_name{2014-12-05-11-09-10}
\def\img_format{}
\centering
\begin{subfigure}{\subfig_width\textwidth}
	\includegraphics[width=1\textwidth]{images/pvnet-select_crop/extend/pose/posenet-\seq_name-clip_t-50\img_format}
\end{subfigure}%
\begin{subfigure}{\subfig_width\textwidth}
	\includegraphics[width=1\textwidth]{images/pvnet-select_crop/extend/pose/mapnet-\seq_name-clip_t-50\img_format}
\end{subfigure}%
\begin{subfigure}{\subfig_width\textwidth}
	\includegraphics[width=1\textwidth]{images/pvnet-select_crop/extend/pose/pvnet_e58_v0-\seq_name-clip_t-50\img_format}
\end{subfigure}%
\begin{subfigure}{\subfig_width\textwidth}
	\includegraphics[width=1\textwidth]{images/pvnet-select_crop/extend/cdf/\seq_name_trans_cdf.pdf}
\end{subfigure}%

\begin{subfigure}{\subfig_width\textwidth}
	\includegraphics[width=1\textwidth]{images/pvnet-select_crop/extend/pose/posenet-\seq_name-clip_t-100\img_format}
	\centerline{PoseNet \cite{kendall2015posenet, kendall2016modelling, kendall2017geometric}}
\end{subfigure}%
\begin{subfigure}{\subfig_width\textwidth}
	\includegraphics[width=1\textwidth]{images/pvnet-select_crop/extend/pose/mapnet-\seq_name-clip_t-100\img_format}
	\centerline{MapNet \cite{brahmbhatt2018mapnet}}
\end{subfigure}%
\begin{subfigure}{\subfig_width\textwidth}
	\includegraphics[width=1\textwidth]{images/pvnet-select_crop/extend/pose/pvnet_e58_v0-\seq_name-clip_t-100\img_format}
	\centerline{Ours}
\end{subfigure}%
\begin{subfigure}{\subfig_width\textwidth}
	\includegraphics[width=1\textwidth]{images/pvnet-select_crop/extend/cdf/\seq_name_rot_cdf.pdf}
	\centerline{Cumulative errors}
\end{subfigure}%

\caption{Results of PoseNet, MapNet and our model on the FULL3 scenes of the Oxford RobotCar dataset \cite{robotcar2017}. The red and black lines indicate predicted and ground truth poses respectively. The star represents the start point. The poses of PoseNet and MapNet are from \cite{brahmbhatt2018mapnet}. To better visualize the trajectories, we select points with translation errors within 50m (top) and 100m (bottom). Cumulative translation (top right) and rotation (bottom right) errors are also illustrated.}
	\label{fig:exp_robotcar_110910}
\end{figure*}

Fig.~\ref{fig:exp_robotcar_110910}, \ref{fig:exp_robotcar_091832}, \ref{fig:exp_robotcar_084510}, \ref{fig:exp_robotcar_123219} show the trajectories of PoseNet, MapNet and our model. To better visualize the comparison against PoseNet and MapNet, we plot the estimated poses of frames with translation errors within the range of 50m and 100m, respectively. Cumulative translation and rotation distribution errors of three models are illustrated as well. As can be seen obviously, our method gives much more accurate translation and rotation predictions, especially in scenes where PoseNet and MapNet produce lots of outliers.

\subsection{Failure Cases}
As shown in Fig.~\ref{fig:exp_robotcar_181050} and \ref{fig:exp_robotcar_181843}, all the three methods behave poorly in sequences captured at night (FULL7 and FULL8), although our methods gives better performance. The major reason is that pixel values are changed too much between day and night. The performance can be improved by fine-tuning on data with similar conditions. Semantic information can also be introduced in the future.

\section*{Feature Visualization}
\begin{figure*}[p]
	\def\subfig_width{0.2}
\centering
\begin{subfigure}{\subfig_width\textwidth}
	\centering
	\includegraphics[width=0.95\textwidth]{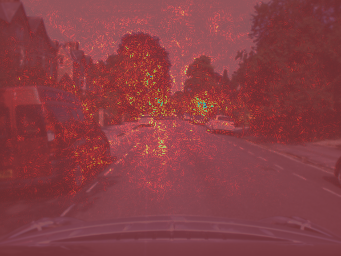}
\end{subfigure}%
\begin{subfigure}{\subfig_width\textwidth}
	\includegraphics[width=0.95\textwidth]{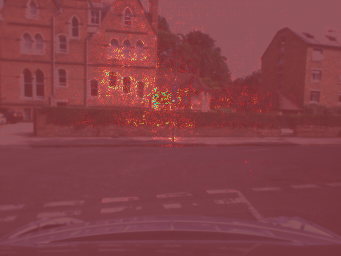}
\end{subfigure}%
\begin{subfigure}{\subfig_width\textwidth}
	\includegraphics[width=0.95\textwidth]{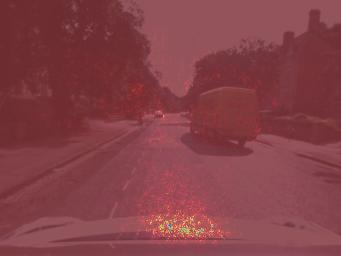}
\end{subfigure}%
\begin{subfigure}{\subfig_width\textwidth}
	\includegraphics[width=0.95\textwidth]{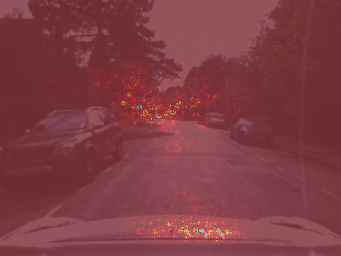}
\end{subfigure}%
\begin{subfigure}{\subfig_width\textwidth}
	\includegraphics[width=0.95\textwidth]{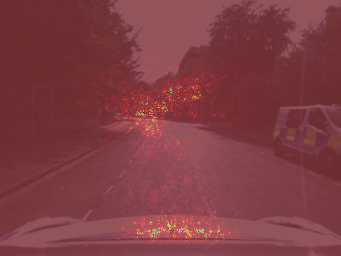}
\end{subfigure}%
\vspace{0.2cm}

\begin{subfigure}{\subfig_width\textwidth}
	\centering
	\includegraphics[width=0.95\textwidth]{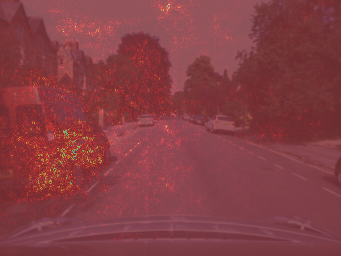}
\end{subfigure}%
\begin{subfigure}{\subfig_width\textwidth}
	\includegraphics[width=0.95\textwidth]{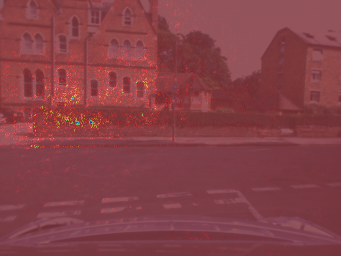}
\end{subfigure}%
\begin{subfigure}{\subfig_width\textwidth}
	\includegraphics[width=0.95\textwidth]{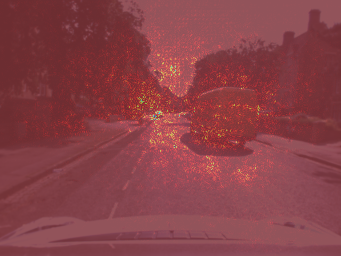}
\end{subfigure}%
\begin{subfigure}{\subfig_width\textwidth}
	\includegraphics[width=0.95\textwidth]{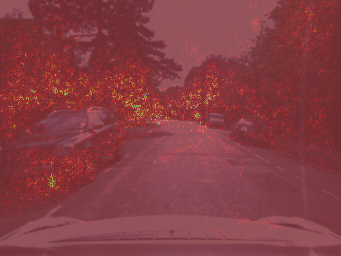}
\end{subfigure}%
\begin{subfigure}{\subfig_width\textwidth}
	\includegraphics[width=0.95\textwidth]{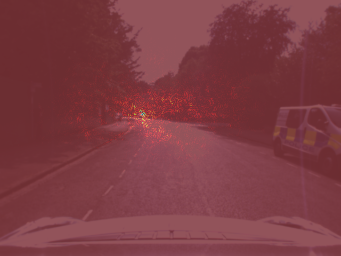}
\end{subfigure}%
\vspace{0.2cm}
\begin{subfigure}{\subfig_width\textwidth}
	\centering
	\includegraphics[width=0.95\textwidth]{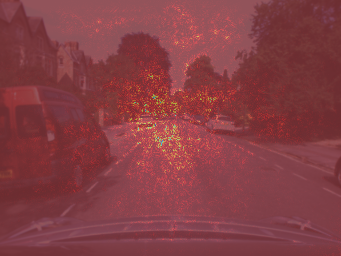}
	\subcaption{}
	\label{fig:vis_a}
\end{subfigure}%
\begin{subfigure}{\subfig_width\textwidth}
	\includegraphics[width=0.95\textwidth]{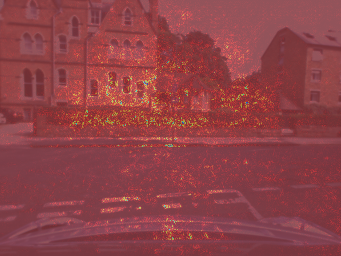}
	\subcaption{}
	\label{fig:vis_b}
\end{subfigure}%
\begin{subfigure}{\subfig_width\textwidth}
	\includegraphics[width=0.95\textwidth]{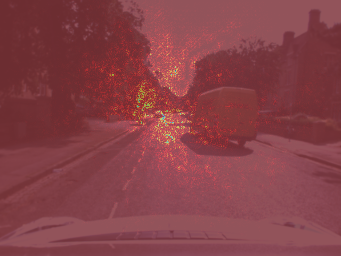}
	\subcaption{}
	\label{fig:vis_c}
\end{subfigure}%
\begin{subfigure}{\subfig_width\textwidth}
	\includegraphics[width=0.95\textwidth]{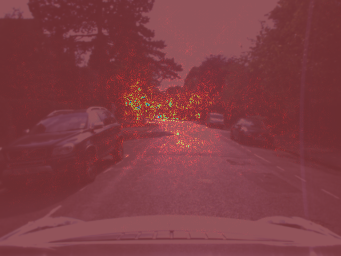}
	\subcaption{}
	\label{fig:vis_d}
\end{subfigure}%
\begin{subfigure}{\subfig_width\textwidth}
	\includegraphics[width=0.95\textwidth]{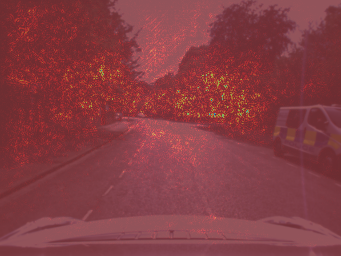}
	\subcaption{}
	\label{fig:vis_e}
\end{subfigure}%

\caption{Attention maps of example images for PoseNet \cite{kendall2015posenet, kendall2016modelling, kendall2017geometric} (top), MapNet \cite{brahmbhatt2018mapnet} (middle) and our model (bottom) on the Oxford RobotCar dataset \cite{robotcar2017}. Compared with PoseNet and MapNet, our model focuses more on static objects and regions with geometric meanings. Both local and global are concentrated in our method to mitigate local similar appearances.}
\label{fig:vis_samples}
\end{figure*}

In Fig.~\ref{fig:vis_samples}, we visualize the attention maps of images for PoseNet, MapNet and our model. We observe that PoseNet and MapNet rely heavily on local regions including dynamic objects such cars (Fig.~\ref{fig:vis_a}, \ref{fig:vis_c}, \ref{fig:vis_d}). Even the front part of the moving car, which is used for data collection, is covered in the salient maps produced by PoseNet (Fig.~\ref{fig:vis_c}, \ref{fig:vis_d}, \ref{fig:vis_e}). These regions are either easily changed over time or sensitive to similar appearances, leading to severe localization uncertainties. In contrast, our model emphasizes on stable features such as buildings (Fig.~\ref{fig:vis_b}) and roads (Fig.~\ref{fig:vis_c}, \ref{fig:vis_d}). Moreover, both local and global regions (Fig.~\ref{fig:vis_b}, \ref{fig:vis_e}) are included. Benefited from the content augmentation, unstable features are suppressed while robust features are advocated.

\label{sec:visualization}

\setlength{\tabcolsep}{1.4pt}

\begin{figure*}[t]
	
\def\subfig_width{0.25}
\def\seq_name{2014-11-25-09-18-32}
\centering
\begin{subfigure}{\subfig_width\textwidth}
	\includegraphics[width=1\textwidth]{images/pvnet-select_crop/extend/pose/posenet-\seq_name-clip_t-50}
\end{subfigure}%
\begin{subfigure}{\subfig_width\textwidth}
	\includegraphics[width=1\textwidth]{images/pvnet-select_crop/extend/pose/mapnet-\seq_name-clip_t-50}
\end{subfigure}%
\begin{subfigure}{\subfig_width\textwidth}
	\includegraphics[width=1\textwidth]{images/pvnet-select_crop/extend/pose/pvnet_e58_v0-\seq_name-clip_t-50}
\end{subfigure}%
\begin{subfigure}{\subfig_width\textwidth}
	\includegraphics[width=1\textwidth]{images/pvnet-select_crop/extend/cdf/\seq_name_trans_cdf}
\end{subfigure}%

\begin{subfigure}{\subfig_width\textwidth}
	\includegraphics[width=1\textwidth]{images/pvnet-select_crop/extend/pose/posenet-\seq_name-clip_t-100}
	\centerline{PoseNet \cite{kendall2015posenet, kendall2016modelling, kendall2017geometric}}
\end{subfigure}%
\begin{subfigure}{\subfig_width\textwidth}
	\includegraphics[width=1\textwidth]{images/pvnet-select_crop/extend/pose/mapnet-\seq_name-clip_t-100}
	\centerline{MapNet \cite{brahmbhatt2018mapnet}}
\end{subfigure}%
\begin{subfigure}{\subfig_width\textwidth}
	\includegraphics[width=1\textwidth]{images/pvnet-select_crop/extend/pose/pvnet_e58_v0-\seq_name-clip_t-100}
	\centerline{Ours}
\end{subfigure}%
\begin{subfigure}{\subfig_width\textwidth}
	\includegraphics[width=1\textwidth]{images/pvnet-select_crop/extend/cdf/\seq_name_rot_cdf}
	\centerline{Cumulative errors}
\end{subfigure}%
	
\caption{Results of PoseNet, MapNet and our model on the FULL4 scenes of the Oxford RobotCar dataset \cite{robotcar2017}. The red and black lines indicate predicted and ground truth poses respectively. The star represents the start point. The poses of PoseNet and MapNet are from \cite{brahmbhatt2018mapnet}. To better visualize the trajectories, we select points with translation errors within 50m (top) and 100m (bottom). Cumulative translation (top right) and rotation (bottom right) errors are also illustrated.}
	\label{fig:exp_robotcar_091832}
\end{figure*}
\setlength{\tabcolsep}{1.4pt}

\begin{figure*}[t]
	
\def\subfig_width{0.25}
\def\seq_name{2015-02-03-08-45-10}
\centering
\begin{subfigure}{\subfig_width\textwidth}
	\includegraphics[width=1\textwidth]{images/pvnet-select_crop/extend/pose/posenet-\seq_name-clip_t-50}
\end{subfigure}%
\begin{subfigure}{\subfig_width\textwidth}
	\includegraphics[width=1\textwidth]{images/pvnet-select_crop/extend/pose/mapnet-\seq_name-clip_t-50}
\end{subfigure}%
\begin{subfigure}{\subfig_width\textwidth}
	\includegraphics[width=1\textwidth]{images/pvnet-select_crop/extend/pose/pvnet_e58_v0-\seq_name-clip_t-50}
\end{subfigure}%
\begin{subfigure}{\subfig_width\textwidth}
	\includegraphics[width=1\textwidth]{images/pvnet-select_crop/extend/cdf/\seq_name_trans_cdf}
\end{subfigure}%

\begin{subfigure}{\subfig_width\textwidth}
	\includegraphics[width=1\textwidth]{images/pvnet-select_crop/extend/pose/posenet-\seq_name-clip_t-100}
	\centerline{PoseNet \cite{kendall2015posenet, kendall2016modelling, kendall2017geometric}}
\end{subfigure}%
\begin{subfigure}{\subfig_width\textwidth}
	\includegraphics[width=1\textwidth]{images/pvnet-select_crop/extend/pose/mapnet-\seq_name-clip_t-100}
	\centerline{MapNet \cite{brahmbhatt2018mapnet}}
\end{subfigure}%
\begin{subfigure}{\subfig_width\textwidth}
	\includegraphics[width=1\textwidth]{images/pvnet-select_crop/extend/pose/pvnet_e58_v0-\seq_name-clip_t-100}
	\centerline{Ours}
\end{subfigure}%
\begin{subfigure}{\subfig_width\textwidth}
	\includegraphics[width=1\textwidth]{images/pvnet-select_crop/extend/cdf/\seq_name_rot_cdf}
	\centerline{Cumulative errors}
\end{subfigure}%

\caption{Results of PoseNet, MapNet and our model on the FULL5 scenes of the Oxford RobotCar dataset \cite{robotcar2017}. The red and black lines indicate predicted and ground truth poses respectively. The star represents the start point. The poses of PoseNet and MapNet are from \cite{brahmbhatt2018mapnet}. To better visualize the trajectories, we select points with translation errors within 50m (top) and 100m (bottom). Cumulative translation (top right) and rotation (bottom right) errors are also illustrated.}
	\label{fig:exp_robotcar_084510}
\end{figure*}
\setlength{\tabcolsep}{1.4pt}

\begin{figure*}[t]
	
\def\subfig_width{0.25}
\def\seq_name{2015-02-24-12-32-19}
\centering
\begin{subfigure}{\subfig_width\textwidth}
	\includegraphics[width=1\textwidth]{images/pvnet-select_crop/extend/pose/posenet-\seq_name-clip_t-50}
\end{subfigure}%
\begin{subfigure}{\subfig_width\textwidth}
	\includegraphics[width=1\textwidth]{images/pvnet-select_crop/extend/pose/mapnet-\seq_name-clip_t-50}
\end{subfigure}%
\begin{subfigure}{\subfig_width\textwidth}
	\includegraphics[width=1\textwidth]{images/pvnet-select_crop/extend/pose/pvnet_e58_v0-\seq_name-clip_t-50}
\end{subfigure}%
\begin{subfigure}{\subfig_width\textwidth}
	\includegraphics[width=1\textwidth]{images/pvnet-select_crop/extend/cdf/\seq_name_trans_cdf}
\end{subfigure}%

\begin{subfigure}{\subfig_width\textwidth}
	\includegraphics[width=1\textwidth]{images/pvnet-select_crop/extend/pose/posenet-\seq_name-clip_t-100}
	\centerline{PoseNet \cite{kendall2015posenet, kendall2016modelling, kendall2017geometric}}
\end{subfigure}%
\begin{subfigure}{\subfig_width\textwidth}
	\includegraphics[width=1\textwidth]{images/pvnet-select_crop/extend/pose/mapnet-\seq_name-clip_t-100}
	\centerline{MapNet \cite{brahmbhatt2018mapnet}}
\end{subfigure}%
\begin{subfigure}{\subfig_width\textwidth}
	\includegraphics[width=1\textwidth]{images/pvnet-select_crop/extend/pose/pvnet_e58_v0-\seq_name-clip_t-100}
	\centerline{Ours}
\end{subfigure}%
\begin{subfigure}{\subfig_width\textwidth}
	\includegraphics[width=1\textwidth]{images/pvnet-select_crop/extend/cdf/\seq_name_rot_cdf}
	\centerline{Cumulative errors}
\end{subfigure}%

\caption{Results of PoseNet, MapNet and our model on the FULL6 scenes of the Oxford RobotCar dataset \cite{robotcar2017}. The red and black lines indicate predicted and ground truth poses respectively. The star represents the start point. The poses of PoseNet and MapNet are from \cite{brahmbhatt2018mapnet}. To better visualize the trajectories, we select points with translation errors within 50m (top) and 100m (bottom). Cumulative translation (top right) and rotation (bottom right) errors are also illustrated.}
	\label{fig:exp_robotcar_123219}
\end{figure*}
\setlength{\tabcolsep}{1.4pt}

\begin{figure*}[t]
	
\def\subfig_width{0.25}
\def\seq_name{2014-12-10-18-10-50}
\centering
\begin{subfigure}{\subfig_width\textwidth}
	\includegraphics[width=1\textwidth]{images/pvnet-select_crop/extend/pose/posenet-\seq_name-clip_t-50}
\end{subfigure}%
\begin{subfigure}{\subfig_width\textwidth}
	\includegraphics[width=1\textwidth]{images/pvnet-select_crop/extend/pose/mapnet-\seq_name-clip_t-50}
\end{subfigure}%
\begin{subfigure}{\subfig_width\textwidth}
	\includegraphics[width=1\textwidth]{images/pvnet-select_crop/extend/pose/pvnet_e58_v0-\seq_name-clip_t-50}
\end{subfigure}%
\begin{subfigure}{\subfig_width\textwidth}
	\includegraphics[width=1\textwidth]{images/pvnet-select_crop/extend/cdf/\seq_name_trans_cdf}
\end{subfigure}%

\begin{subfigure}{\subfig_width\textwidth}
	\includegraphics[width=1\textwidth]{images/pvnet-select_crop/extend/pose/posenet-\seq_name-clip_t-100}
	\centerline{PoseNet \cite{kendall2015posenet, kendall2016modelling, kendall2017geometric}}
\end{subfigure}%
\begin{subfigure}{\subfig_width\textwidth}
	\includegraphics[width=1\textwidth]{images/pvnet-select_crop/extend/pose/mapnet-\seq_name-clip_t-100}
	\centerline{MapNet \cite{brahmbhatt2018mapnet}}
\end{subfigure}%
\begin{subfigure}{\subfig_width\textwidth}
	\includegraphics[width=1\textwidth]{images/pvnet-select_crop/extend/pose/pvnet_e58_v0-\seq_name-clip_t-100}
	\centerline{Ours}
\end{subfigure}%
\begin{subfigure}{\subfig_width\textwidth}
	\includegraphics[width=1\textwidth]{images/pvnet-select_crop/extend/cdf/\seq_name_rot_cdf}
	\centerline{Cumulative errors}
\end{subfigure}%

\caption{Results of PoseNet, MapNet and our model on the FULL7 scenes of the Oxford RobotCar dataset \cite{robotcar2017}. The red and black lines indicate predicted and ground truth poses respectively. The star represents the start point. The poses of PoseNet and MapNet are from \cite{brahmbhatt2018mapnet}. To better visualize the trajectories, we select points with translation errors within 50m (top) and 100m (bottom). Cumulative translation (top right) and rotation (bottom right) errors are also illustrated.}
	\label{fig:exp_robotcar_181050}
\end{figure*}
\setlength{\tabcolsep}{1.4pt}

\begin{figure*}[t]
\def\subfig_width{0.25}
\def\seq_name{2014-12-17-18-18-43}
\centering
\begin{subfigure}{\subfig_width\textwidth}
	\includegraphics[width=1\textwidth]{images/pvnet-select_crop/extend/pose/posenet-\seq_name-clip_t-50}
\end{subfigure}%
\begin{subfigure}{\subfig_width\textwidth}
	\includegraphics[width=1\textwidth]{images/pvnet-select_crop/extend/pose/mapnet-\seq_name-clip_t-50}
\end{subfigure}%
\begin{subfigure}{\subfig_width\textwidth}
	\includegraphics[width=1\textwidth]{images/pvnet-select_crop/extend/pose/pvnet_e58_v0-\seq_name-clip_t-50}
\end{subfigure}%
\begin{subfigure}{\subfig_width\textwidth}
	\includegraphics[width=1\textwidth]{images/pvnet-select_crop/extend/cdf/\seq_name_trans_cdf}
\end{subfigure}%

\begin{subfigure}{\subfig_width\textwidth}
	\includegraphics[width=1\textwidth]{images/pvnet-select_crop/extend/pose/posenet-\seq_name-clip_t-100}
	\centerline{PoseNet \cite{kendall2015posenet, kendall2016modelling, kendall2017geometric}}
\end{subfigure}%
\begin{subfigure}{\subfig_width\textwidth}
	\includegraphics[width=1\textwidth]{images/pvnet-select_crop/extend/pose/mapnet-\seq_name-clip_t-100}
	\centerline{MapNet \cite{brahmbhatt2018mapnet}}
\end{subfigure}%
\begin{subfigure}{\subfig_width\textwidth}
	\includegraphics[width=1\textwidth]{images/pvnet-select_crop/extend/pose/pvnet_e58_v0-\seq_name-clip_t-100}
	\centerline{Ours}
\end{subfigure}%
\begin{subfigure}{\subfig_width\textwidth}
	\includegraphics[width=1\textwidth]{images/pvnet-select_crop/extend/cdf/\seq_name_rot_cdf}
	\centerline{Cumulative errors}
\end{subfigure}
\caption{Results of PoseNet, MapNet and our model on the FULL8 scenes of the Oxford RobotCar dataset \cite{robotcar2017}. The red and black lines indicate predicted and ground truth poses respectively. The star represents the start point. The poses of PoseNet and MapNet are from \cite{brahmbhatt2018mapnet}. To better visualize the trajectories, we select points with translation errors within 50m (top) and 100m (bottom). Cumulative translation (top right) and rotation (bottom right) errors are also illustrated.}
	\label{fig:exp_robotcar_181843}
\end{figure*}
\setlength{\tabcolsep}{1.4pt}

\setlength{\tabcolsep}{1.4pt}



\end{document}